%% file: arxiv_v2.tex
\documentclass{article}

\usepackage{microtype}
\usepackage{graphicx}
\usepackage{subfigure}
\usepackage{booktabs} 
\usepackage{subcaption}
\usepackage[utf8]{inputenc} 
\usepackage[T1]{fontenc}    
\usepackage{hyperref}       
\usepackage{url}            
\usepackage{booktabs}       
\usepackage{amsfonts}       
\usepackage{nicefrac}       
\usepackage{microtype}      
\usepackage{multicol}
\usepackage{multirow}
\usepackage{caption}
\usepackage{array}       
\usepackage{wrapfig}
\usepackage{float}
\usepackage{amsmath}
\usepackage{amssymb}
\usepackage{mathtools}
\usepackage{amsthm}
\usepackage{bbding}
\usepackage{enumitem}
\usepackage{adjustbox}  
\usepackage{dblfloatfix}
\usepackage{amssymb}
\usepackage{minitoc}
\usepackage{natbib}

\usepackage[accepted]{mlsys2024}

\DeclareTextFontCommand{\emph}{\em}

\renewcommand{\paragraph}[1]{\noindent\vspace{0.2em}\textbf{{#1}}}

\mlsystitlerunning{QUIK: Towards End-to-end 4-Bit Inference on  Generative Large Language Models}

\begin{document}

\twocolumn[
\mlsystitle{QUIK: Towards End-to-end 4-Bit Inference \\ on  Generative Large Language Models}

\mlsyssetsymbol{equal}{*}

\begin{mlsysauthorlist}
\mlsysauthor{Saleh Ashkboos}{equal,ethz}
\mlsysauthor{Ilia Markov}{equal,ist}
\mlsysauthor{Elias Frantar}{ist}
\mlsysauthor{Tingxuan Zhong}{xidian}
\mlsysauthor{Xingchen Wang}{xidian}
\mlsysauthor{Jie Ren}{kaust}
\mlsysauthor{Torsten Hoefler}{ethz}
\mlsysauthor{Dan Alistarh}{ist,nm}
\end{mlsysauthorlist}

\mlsysaffiliation{ethz}{ETH Zurich}
\mlsysaffiliation{ist}{Institute of Science and Technology Austria}
\mlsysaffiliation{kaust}{KAUST}
\mlsysaffiliation{xidian}{Xidian University}
\mlsysaffiliation{nm}{Neural Magic, Inc.}

\mlsyscorrespondingauthor{Saleh Ashkboos}{saleh.ashkboos@inf.ethz.ch}
\mlsyscorrespondingauthor{Dan Alistarh}{dan.alistarh@ist.ac.at}

\vskip 0.3in

\begin{abstract}
Large Language Models (LLMs) from the GPT family have become extremely popular, leading to a race towards reducing their inference costs to allow for efficient local computation. Yet, the vast majority of existing work focuses on weight-only quantization, which can reduce runtime costs in the memory-bound one-token-at-a-time generative setting, but does not address them in compute-bound scenarios, such as batched inference or prompt processing.  
In this paper, we address the general quantization problem, where both weights and activations should be quantized. We show, for the first time, that the majority of  inference computations for large generative models such as LLaMA, OPT, and Falcon can be performed with both weights and activations being cast to 4 bits, in a way that leads to practical speedups, while at the same time maintaining good accuracy. We achieve this via a hybrid quantization strategy called QUIK, which compresses most of the weights and activations to 4-bit, while keeping some outlier weights and activations in higher-precision. The key feature of our scheme is that it is designed with computational efficiency in mind: we provide GPU kernels matching the QUIK format with highly-efficient layer-wise runtimes, which lead to practical end-to-end throughput improvements of up to 3.4x relative to FP16 execution. 
 We provide detailed studies for models from the OPT, LLaMA-2 and Falcon families, as well as a first instance of accurate inference using quantization plus 2:4 sparsity.  
Code is available at: \url{https://github.com/IST-DASLab/QUIK}.
\end{abstract}
]

\printAffiliationsAndNotice{\mlsysEqualContribution} 

\vspace{-1em}
\section{Introduction}
\label{sec:intro}
\input{intro.tex}

\section{Motivation}

\paragraph{Roofline Analysis.} 
To motivate our focus on the compute-bound case,  we begin an analysis of the basic computational operation in the context of LLMs, a matrix multiplication for different numbers of tokens. We profile a linear layer of standard size (11K x 4K, corresponding to the MLP in LLaMA-7B~\cite{touvron2023llama2}), using the NVIDIA NSight Compute toolkit~\cite{nvidia_nsight}, from a single token to 16, 256 and 1024 tokens. 

\begin{figure}[t!]
\centering
\includegraphics[width=0.4\textwidth]{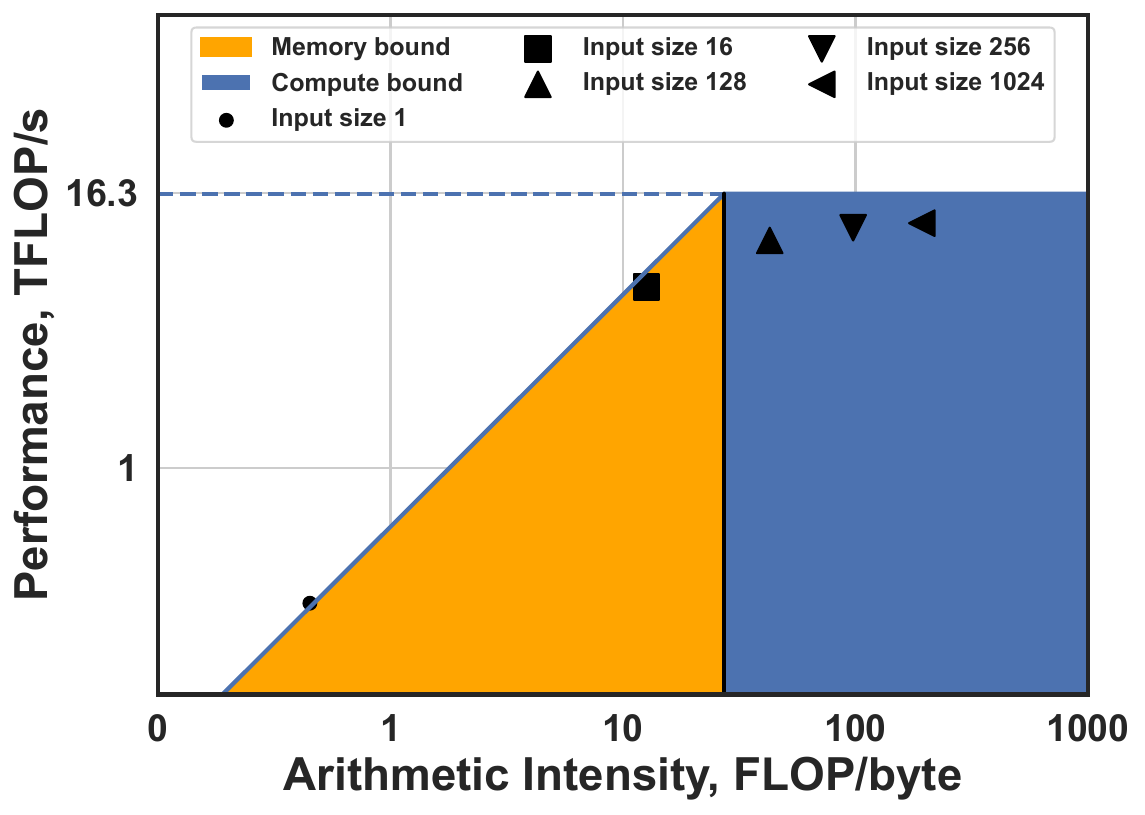}
\caption{Roofline analysis of a standard LLM MatMul operation, for a matrix of size 8K x 8K, in FP32,  on an NVIDIA GPU. Markers denote the results of profiling with different token counts (from 1 to 1024). 
Small counts (1 and 16) are memory-bound, whereas larger counts (from 128 to 1024) are compute-bound.}
\label{fig:roofline}
\vspace{-.5em}
\end{figure}

The results, illustrated in Figure~\ref{fig:roofline}, clearly show that the case of few tokens (1 and 16) is memory-bound, whereas the workload becomes compute-bound for the larger token counts, specifically larger than 128. A realistic end-to-end LLM deployment would need to consider optimizing both scenarios, as the prompt processing ``prefill'' case falls into the large token count scenario, whereas generating one-token-at-a-time falls into the former case. 
Moreover, running a ``batched'' version of the single-token workload, i.e. for multiple users, would again result in large token counts, returning to the compute-bound case. 

Further, we observe that existing methods for weight-only quantization, e.g.~\cite{frantar2022gptq, dettmers2022case, lin2023awq} only serve to improve the arithmetic intensity of this operation, by reducing the amount of data which needs to be transferred per operation, but still perform the computation in the original precision. Thus, they do not help in the compute-bound case, and in fact even \emph{slightly increase} the amount of computation per operation, due to the de-quantization overheads.

\begin{figure}[t]
\centering
\includegraphics[width=0.45\textwidth]{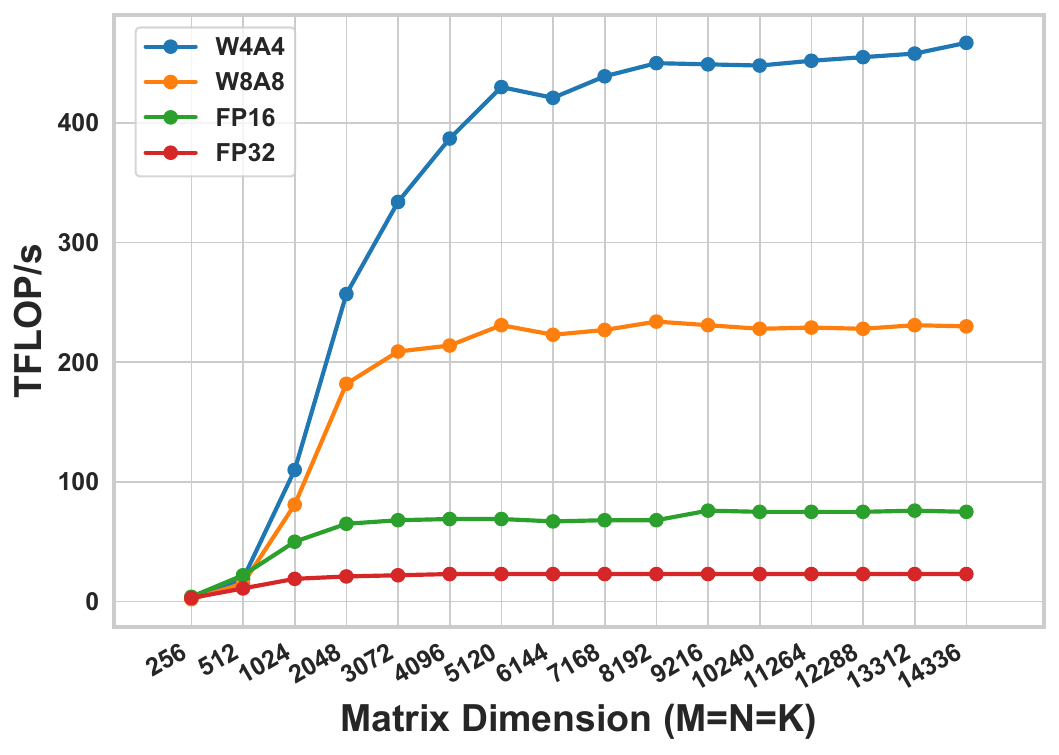}
\caption{\label{fig:ideal-speedups} Ideal matrix multiplication performance for different layer sizes and data precision on RTX3090.}
\end{figure}

\paragraph{Speedup Potential.} 
Given our focus on the compute-bound case, it is natural to investigate the available hardware options leading to potential speedups. 
As shown in Figure~\ref{fig:ideal-speedups}, quantization is a natural approach in this case, given that NVIDIA GPUs have native support for INT4 and INT8 data types, providing major throughput improvements across matrix sizes. Specifically, INT8 provides throughput improvements that can be slightly higher than 2x relative to FP16 on raw MatMuls, whereas INT4 almost doubles over INT8. However, to leverage these hardware operations, \emph{both layer inputs (activations) and layer weights} must be quantized to the same compressed data type. 

We will focus on accurate post-training quantization for LLM inference, by compressing both weights and activations, primarily to INT4 data types. 
As stated, weight-only quantization~\cite{frantar2022gptq, lin2023awq} does not transfer to our setting, and activation quantization is notoriously challenging~\cite{xiao2022smoothquant}. Moreover, as shown in Table~\ref{tab:accuracy_results_opt}, existing methods for quantizing both weights and activations in LLMs break down in terms of accuracy when applied to 4bit compression.

\section{Method}
\label{sec:method}
\input{QUIK.tex}

\vspace{-0.4em}
\section{Experimental Validation}
\label{sec:experiments}
\input{experiments.tex}

\vspace{-0.5em}
\section{Conclusion and Future Work}

\label{sec:conclusion}
\input{conclusion.tex}

\nocite{langley00}

\bibliography{References}
\bibliographystyle{mlsys2024}


\appendix
\label{sec:appendix}
\input{appendix.tex}

%


\end{document}

%% file: intro.tex
Large language models (LLMs) from the Generative Pretrained Transformer (GPT) family~\cite{radford2019language} are massively popular. One key contributor to their adoption has been the ability to compress them using advanced techniques, e.g.,~\cite{frantar2022gptq, dettmers2022llm, lin2023awq, yuan2023rptq}, enabling local storage and efficient generative inference for these models, even on personal computers.  
The vast majority of work on LLM quantization can be categorized into two cases:
\begin{itemize}[leftmargin=*]
    \item \emph{Weight-only quantization methods}~\cite{frantar2022gptq, dettmers2022llm, lin2023awq, dettmers2023spqr, lin2023awq, kim2023squeezellm} that help reduce the massive memory-transfer costs of LLM inference. Yet, these methods do not reduce computation, and cannot provide significant speedup for computationally-bound settings, such as prompt processing or 
 batch inference.
    \item \emph{Joint weight-activation quantization methods}, which can provide computational improvements, but either focus exclusively on 8-bit weights and activations (8W8A)~\cite{xiao2022smoothquant, dettmers2022llm}, or execute with large amounts of accuracy loss relative to their uncompressed counterparts~\cite{yuan2023rptq, shao2023omniquant}.
\end{itemize}

Thus, there is still a significant gap between compressed formats efficiently supported by hardware---specifically, NVIDIA GPUs natively support accelerated 4bit matrix multiplication on both the Ampere and Lovelace architectures~\cite{nvidia_cutlass}---and quantization algorithms with computational support which would allow inference to be performed accurately on such compressed formats. 

\paragraph{Contribution.} In this paper, we take a step towards bridging this gap,  and show for the first time that a large fraction of the  computation in modern LLMs such as OPT~\cite{zhang2022opt}, LLaMA-2~\cite{touvron2023llama2} and Falcon~\cite{falcon2023} can be performed accurately and efficiently using \textit{4-bit activations and weights (4W4A)}. 

\begin{figure}[t]
        \centering
        \hspace*{0.5cm}
        \includegraphics[width=0.85\linewidth]{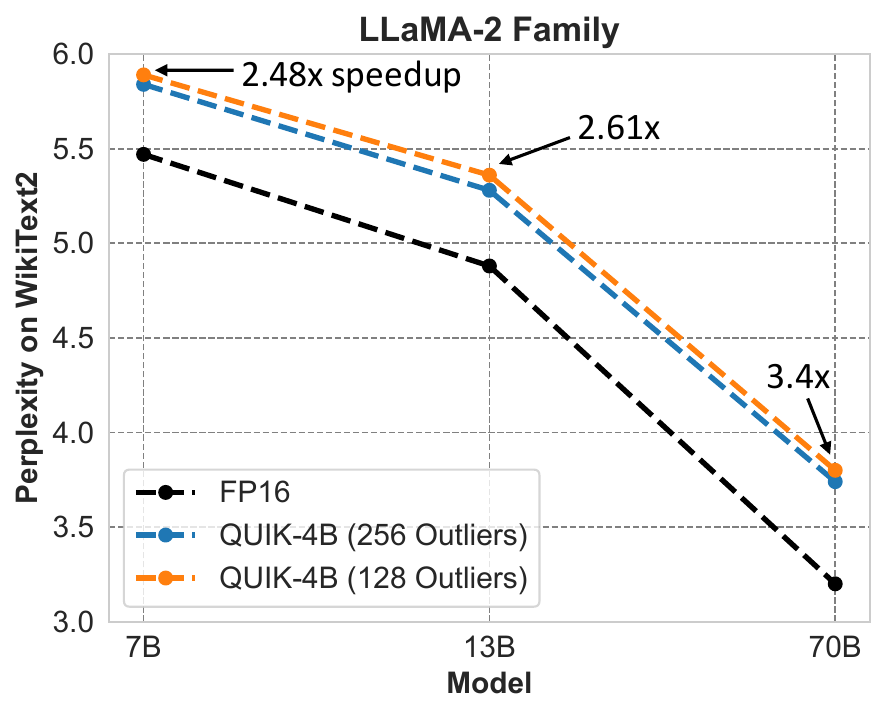}
           \caption{Accuracy and speedups for QUIK at different model sizes, on the LLaMA family of models. QUIK achieves up to 3.4x speedup with minor accuracy degradation on LLaMA-2 models.}
    \label{fig:llama-end-to-end}
    \end{figure}

On the algorithmic side, we show significantly improved results relative to prior work on joint quantization of weights and activations to 4 bits, via a hybrid scheme for \textbf{QU}antization to \textbf{I}NT4 with GPU \textbf{K}ernel support, called \textbf{QUIK}. In QUIK, matrices are split into ``base'' weights and activations, which are processed exclusively at 4-bit precision, and a small number of ``outlier'' weights and activations, which are processed at higher precision such as INT8 or FP16. Using this approach, as well as additional insights into layer sensitivity, 
we build a framework which can recover accuracy within 
0.3--0.5 perplexity points across model sizes (corresponding to 6\%-16\% relative error), while executing a large fraction of the inference in INT4. 
For illustration, for the sensitive LLaMA2  model with 70B parameters, we can recover accuracy within 0.5 perplexity, while executing 70\% of the linear layer computations in INT4, leading to 3.4x end-to-end speedups (see Figure~\ref{fig:llama-end-to-end}). 

On the systems side, 
the key feature of QUIK is that it 
can be implemented efficiently via GPU kernels with low runtime and memory overheads relative to GPU-native INT4 matrix multiplication (MatMul). 
We demonstrate this via a general implementation leading to per-layer speedups and end-to-end throughput improvements relative to both FP16 and INT8 baselines. Specifically, we show that supporting a limited number of feature and weight outliers can have negligible overhead by fusing the quantization and dequantization operations into the MatMul and by mitigating their costs in linear layers via additional optimizations.

Overall, QUIK leverages quantization for significant end-to-end speedups and memory reductions. 
For example, for processing a sequence of 2048 tokens on a commodity RTX 3090 GPU, we achieve end-to-end speedups between 3.1x, for the OPT-66B and Falcon-180B models, and 3.4x for LLaMA2-70B, relative to a theoretical optimum of $\approx$4x. 
In addition, QUIK requires much less GPU memory, and therefore, less GPUs, relative to FP16. For instance, QUIK provides 3.6x memory reduction for OPT-66B, and 3x compression for accurate execution of LLaMA2-70B, executing the latter in less than 50GB of GPU memory.

%% file: QUIK.tex
\subsection{Background}

We focus on the task of accelerating linear layers within Large Language Models (LLMs) by employing 4-bit quantization for both the weight matrix $\mathbf{W}$ and the input matrix $\mathbf{X}$. Following the PyTorch definition \cite{paszke2019pytorch}, a linear layer carries out a linear transformation along with a bias vector $\mathbf{b}$, taking the form of $\mathbf{X} \mathbf{W^T + b}$. We now describe the background and details of the technique.

\paragraph{Outliers in Input Quantization.}
It is known that the activation 
matrices are hard to quantize \cite{dettmers2022llm, xiao2022smoothquant, yuan2023rptq}, mainly due to the presence of \emph{outlier features} in these matrices, where some of the columns have up to 100x larger magnitudes.
LLM.int8() \cite{dettmers2022llm} identifies and extracts the outlier columns of $\mathbf{X}$ during the forward pass and quantizes the rest of the elements with 8-bit. However, LLM.int8() is not efficient at runtime due to the added computational cost of determining outliers on-the-fly. 
Recent work \cite{xiao2022smoothquant} has shown that the outlier features are fixed for each layer across  datasets, which means that {we can extract outlier indices offline} using a small calibration set.

\paragraph{GPTQ Weight Quantization.}
GPTQ~\citep{frantar2022gptq} is a weight-only quantization method which involves the quantization of $\mathbf{W}$ while retaining the activations $\mathbf{X}$ in FP16.
To do this, it iterates over the weight columns; for each column, it quantizes all of its elements simultaneously. Following the quantization of a weight column, GPTQ adjusts the remaining unquantized columns, to the right of the current one, by using second-order information to compensate for the introduced quantization error in the current step.
This process \emph{accumulates the quantization errors at the last columns}, making them more sensitive to quantization.

\begin{figure}[t!]

        \centering
        \includegraphics[width=0.7\columnwidth]{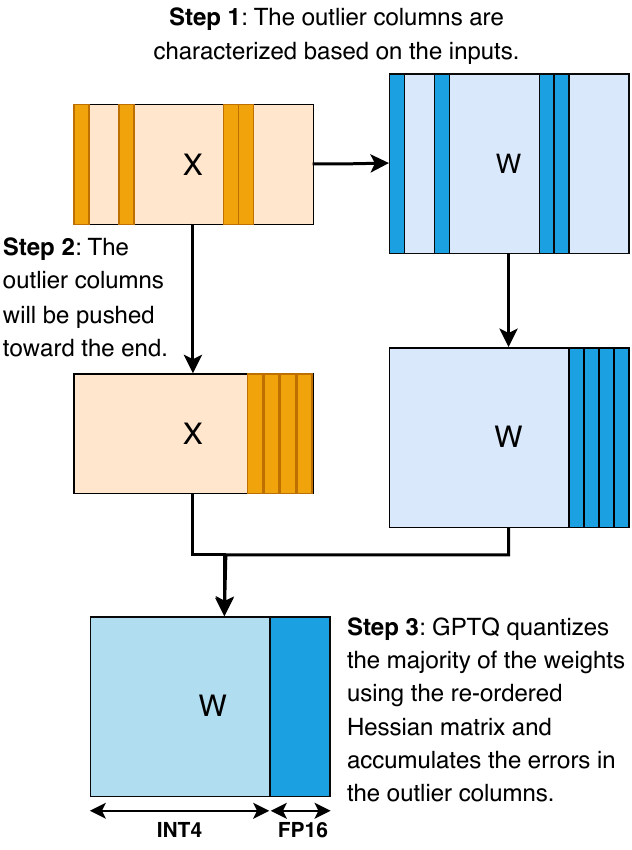}
\caption{Outlier-aware quantization with QUIK. Outlier weight columns are extracted
    based on outlier columns in the input. We permute the outlier columns
    toward the end of the matrix before applying GPTQ quantization (using the re-ordered Hessian matrix) to accumulate the
    quantization errors in the FP16 columns.}     \label{fig:scheme_vis}
\end{figure}

\subsection{QUIK Quantization}
\label{sec:quik_main_desc}
\paragraph{Overview.} At a high level, QUIK 
 works as follows. First, note that, during the  linear transformation $\mathbf{X} \mathbf{W^T}$, the outlier columns in $\mathbf{X}$, by which we mean the columns with large average values defined previously, 
will always be multiplied by certain columns in $\mathbf{W^T}$, as illustrated in Figure \ref{fig:scheme_vis}.
We leverage this observation to improve the quality of GPTQ quantization, in a setting where we  quantize (part of) the activations as well. 

Since the outlier columns are fixed across datasets, we begin by extracting the indices of the outlier columns by means of a calibration set. 
Then,  we rearrange the weight columns (and their corresponding input columns), to shift the outliers  toward the end.
Finally, we perform quantization on the weight columns up to the index of the outliers. This circumvents quantization of these ``difficult'' columns. It also helps GPTQ quantization by 1) aggregating the quantization errors 
to the columns we keep in FP16, and 2) removing potential weight outliers from the 4bit  quantization scale.

\paragraph{Weight Clipping.}
Weight clipping improves quantization by trimming the input distribution before rounding. 
This could be done by either training the whole network to find the optimal clipping thresholds \cite{shao2023omniquant, esser2019learned, choi2018pact}; or employing heuristic methods \cite{lin2023awq, lee2023owq, kim2023squeezellm}.
We found that applying linear search over the clipping thresholds for weight quantization improves final perplexity.

\paragraph{Sensitivity-Based Partial Quantization.}
Accurately selecting outlier columns is key for QUIK. Following~\citep{xiao2022smoothquant, dettmers2022llm}, we select the columns with the largest $\ell_\infty$ norm as outliers. Since finding these columns dynamically at runtime is costly, we follow~\citep{xiao2022smoothquant} in identifying a predefined set of outliers for each layer via a calibration set (see Section~\ref{sec:experiments}), and quantize the weights offline. We use the same outlier indices for extracting the input outlier columns during the forward pass.

This approach is sufficient for accurate  quantization of models such as OPT~\cite{zhang2022opt} (see Section \ref{sec:experiments}). 
However, highly-accurate massive models such as LLaMA2-70B  present a further challenge due to their FeedForward layers, which involve three linear transformations along with element-wise multiplication, as well as the use of the Sigmoid Linear Unit (SiLU) activations. 
Specifically, our $\ell_\infty$ norm analysis illustrated in Figure~\ref{fig:LLaMA70b_var_max}, suggests that the Down$_{\text{proj}}$ layers are much more sensitive to quantization. 
(\citet{li2023fptq} arrived at a similar observation.) 
Thus, we extend our scheme to  recover accuracy
by quantizing the Down$_{\text{proj}}$ layers to 8 bits instead of 4, without other changes to our method. We illustrate the outlier selection procedure in detail in Section~\ref{sec:llama-case-study}. Figure~\ref{fig:llama70b_flops} presents a detailed analysis of the overall FLOP breakdown to various precisions when quantizing the LLaMA2-70B model via QUIK. 

\subsection{Efficient Inference Implementation}
\label{sec:algorithm}

\begin{figure}[t!]
\centering
\includegraphics[width=0.8\columnwidth]{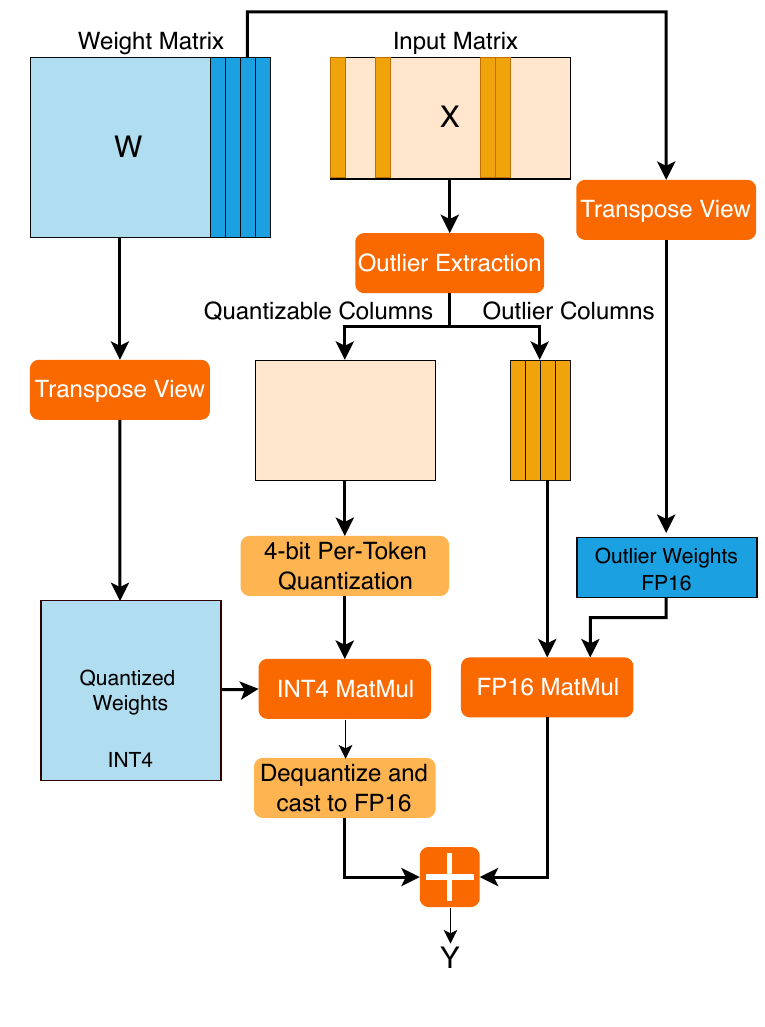}
\vspace{-7mm}
\caption{Schematic for the forward pass of a linear layer ($XW^T$) with QUIK-4B. In the first step, the input outlier features are extracted based on the pre-defined indices and the rest of the input values will be quantized using per-token quantization. The INT4 MatMul will be applied using the quantized weights, calculated offline (see Figure~\ref{fig:scheme_vis}). Finally, the output will be dequantized, cast to FP16, and added to the result of FP16 MatMul.} 
\label{fig:matmul_vis}
\end{figure}

We now provide a high-level description of how models in the QUIK format are executed efficiently on GPU. We illustrate the workflow in Figure~\ref{fig:matmul_vis} and provide pseudocode in Algorithm~\ref{alg:kernels}.
The first and most important step in QUIK is splitting the input matrix of shape (\#tokens, \#features) column-wise, so across features, into two sub-sets, a small  ``full precision'' part (usually half or bfloat16) and a large base part, which will be quantized (see line 3 in the pseudocode). The full-precision part is multiplied with the corresponding (full-precision) part of the weight matrix in standard fashion, while the rest goes through the quantized matrix multiplication pipeline described next.

The quantized MatMul pipeline consists of three parts: 1) dynamically quantizating the activations, 2) actually performing  the MatMul of quantized activations and weights, and 3) dequantizing the result back to  floating point format.

\paragraph{Quantization.} In general, we quantize weights \textit{symmetrically} (only scale)  per output and quantize activations \textit{asymmetrically} (scale and zero) per token. The former is done \textit{offline} (see Section~\ref{sec:quik_main_desc}), while the latter must be done \textit{online} based on the current activation values. Specifically, we first scan the 
 activations to determine the per-token min- and max-value, from which we calculate the scale and zero point (line 12). These are then used to turn the floating point activations into integers,  which are written out again as signed (hence the halfRange subtraction in line 14) INT4 or INT8 values in a packed format for efficient further processing (see lines 13-16).

\paragraph{Matrix Multiplication.} The actual MatMul is performed by the \texttt{CUTLASS} \cite{nvidia_cutlass} library, which is able to effectively utilize the hardware's INT8/INT4 tensor-cores to perform fast low-precision calculations, while accumulating results in a wider INT32 format.

\paragraph{Dequantization.} As the MatMul was carried out purely with quantized INT values, we need to convert back to a floating point format in order to properly integrate scale and zero information. Concretely, we need to multiply each output element $o_{ij}$ by its corresponding input token scale \texttt{scaleAct} and output weight scale \texttt{scaleWeight} (line 22). Additionally, we also need to account for the activation zero-point \texttt{zeroAct}. To do this, we consider a scalar product $\langle w, x \rangle$ 
 (representing a single output value in our overall matmul) where a constant $z$ is added to each $x_i$:
\begin{equation}
    y = \sum_{i} w_i (x_i + z) = \sum_{i} w_i x_i + z \cdot \sum_{i} w_i.
\end{equation}
Consequently, we must shift by $z$ times the \textit{sum over relevant weights}, the latter of which is static and can thus be precomputed as \texttt{wReduced}; the signed to unsigined INT conversion must be considered as well (lines 23 - 24). Finally, we add these dequantized values to the original outlier result, yielding the final output (line 8).

\begin{algorithm}[t]
\caption{Quantization and Dequantization kernels.}
\label{alg:kernels}

\begin{algorithmic}[1]
\FUNCTION{QUIK Matmul}
    \STATE {\bfseries Input:} wInt, wFP, x, FPindices, scaleWeight, wReduced;
    \STATE xFP, xQ $\longleftarrow$ split(x, FPindices); \label{alg_step:split}
    \STATE xINT, zeroAct, scaleAct $\longleftarrow$ Quantization(xQ);\label{alg_step:quantization}
    \STATE resultFP $\longleftarrow$ FPmatmul(xFP, wFP);
    \STATE resultInt $\longleftarrow$ INTmatmul(xInt, wInt);
    \STATE dequantFP $\longleftarrow$ Dequantization(resultInt, zeroAct, scaleAct, scaleWeight, wReduced)
    \STATE \textbf{return} dequantFP + resultFP;
\ENDFUNCTION

\footnotesize
\FUNCTION{Quantization}
    \STATE {\bfseries Input:} dataFP;
    \STATE zeroAct, scaleAct $\longleftarrow$ findZeroScale(dataFP);
    \FOR{elem $\in$ dataFP, outElem $\in$ output}
        \STATE // Use scale/zero corresponding to token
        \STATE outFP $\longleftarrow$ (elem - zeroAct) / scaleAct - halfRange;
        \STATE outElem $\longleftarrow$ pack(outFP);
    \ENDFOR
    \STATE \textbf{return} output, zeroAct, scaleAct;
\ENDFUNCTION

\FUNCTION{Dequantization}
    \STATE {\bfseries Input:} inputINT, zeroAct, scaleAct, scaleWeight, wReduced
    \FOR{elem $\in$ inputINT, outElem $\in$ outputFP}
        \STATE // Use scales for token and weight row, respectively
        \STATE x $\longleftarrow$ elem * scaleAct * scaleWeight;
        \STATE shift $\longleftarrow$ zeroAct + halfRange * scaleAct;
        \STATE shift $\longleftarrow$ shift * wReduced;
        \STATE outElem $\longleftarrow$ x + shift;
    \ENDFOR
    \STATE \textbf{return} outputFP;
\ENDFUNCTION
\end{algorithmic}
\end{algorithm}

\subsection{Performance Optimizations}

The computational backbone of the QUIK kernel implementation is the low-precision \texttt{CUTLASS} matrix multiplication. However, the mixed precision nature of the algorithm imposes the use of auxiliary functions, such as input data splitting, metadata computation, quantization and dequantization. This provides  opportunities for optimizations.

\paragraph{Quantization Fusion.} A naive implementation of the splitting and quantization pipeline would require one read-and-write pass for the outlier-part, another read-and-write pass for the base-part, two read passes to determine per-token min-max values and one more read-and-write pass for actually carrying out quantization. Many of these slow memory-bound operations can be optimized away via careful operator fusion in the form of bespoke kernels.

Specifically, we assign each input row to a CUDA block and perform 3 passes over it: reduction (finding meta information) over the non-outliers elements, quantization of the non-outliers and moving the outliers to a separate piece of memory. This eliminates two costly read (min-max calculation and base-part splitting) and one write pass (base-part splitting), and  overheads of additional kernel launches.

\paragraph{Parallelization Tuning.} For the above quantization procedure to be efficient on a modern GPU, we have to ensure optimal parallelization via careful tuning of CUDA blocks and threadcounts. The most critical tuning parameter is the number of rows we process with one CUDA block. Mapping one block per each row brings additional launching overheads, while mapping too many rows per block results in block over-subscription and lower occupancy of the GPU. Hence, we optimized the appropriate number of rows per block for different matrix sizes (usually values between 8 and 32). This improved quantization speed by up to 30\%.

\paragraph{Dequantization Epilogue.} \texttt{CUTLASS} first accumulates MatMul results in registers before committing them to (slow) global memory. We can avoid an unnecessary write  and read pass of intermediate INT32 matmul results by directly performing dequantization in a custom \textit{epilogue} that is applied before the global memory commit, which we further directly accumulate into the results of the outlier MatMul. Overall, this interleaves two expensive operations and saves additional kernel launches and memory trips.

\paragraph{Performance Impact.} To illustrate the impact of these optimizations, we mark them as different versions of our kernel: version 1 has unfused quantization and dequantization; version 2 has fused quantization and unfused dequantization; version 3  fuses both quantization and dequantization.

\begin{figure}[t!]
\centering
\includegraphics[width=0.4\textwidth]{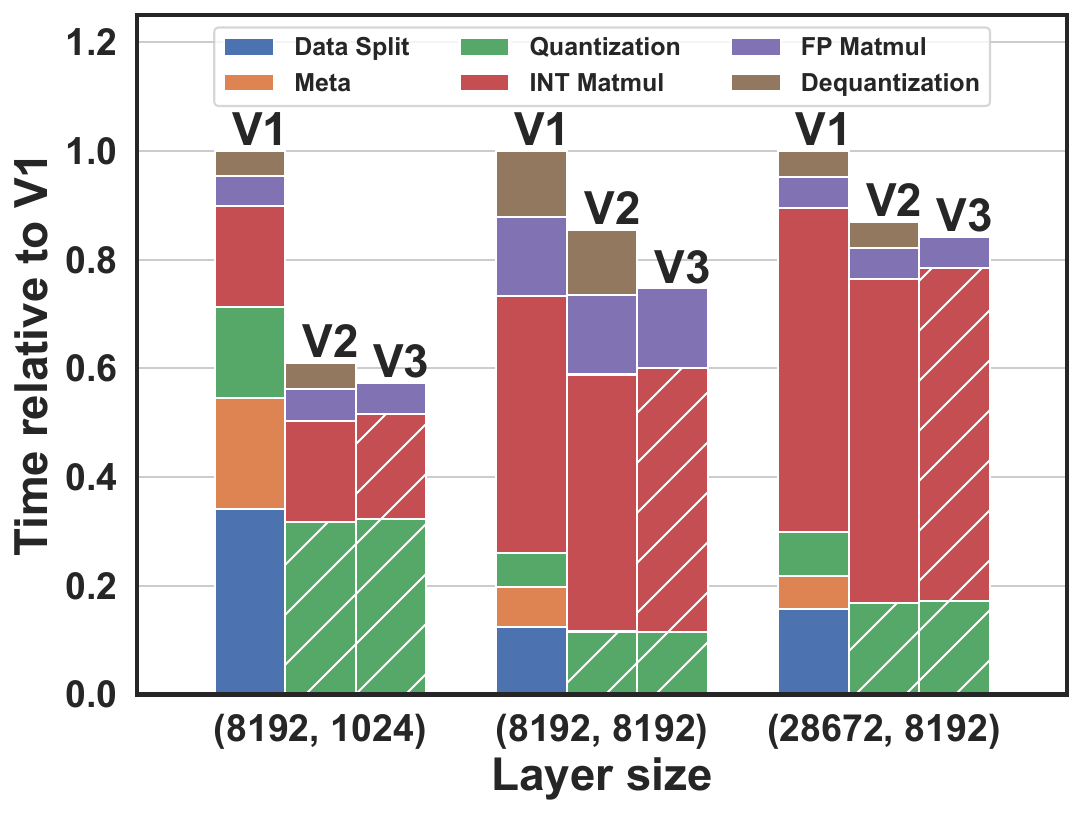}
\vspace{-3mm}
\caption{Operation timings in different QUIK-4B versions with 256 outliers relative to the first version for different matrix sizes. Hatched bars represent fused operations. Experiment executed with input size 2048 on an RTX3090 GPU.}
\label{fig:optim_profile}
\vspace{-2mm}
\end{figure}

Figure~\ref{fig:optim_profile} provides a detailed breakdown of the results of each of these optimizations.  We observe that they are especially effective for the small matrix sizes, where they lead to end-to-end speedups of almost 2x. Fused quantization optimization gives up to 40\% throughput improvement and the dequantization epilogue yields an additional 10\% speedup.

%% file: experiments.tex
\paragraph{General setup.}
We evaluate our method on OPT~\citep{zhang2022opt}, LLaMA-2~\citep{touvron2023llama2}, and Falcon~\citep{falcon2023} models, using HuggingFace~\citep{wolf2019huggingface} implementations of model definitions and datasets. Following SmoothQuant \citep{xiao2022smoothquant}, we extract outlier indices using 512 random sentences from the Pile dataset~\citep{gao2020pile}. We consider up to 5\% (based on the model size) of the input features as outliers in the linear layers. During the GPTQ weight quantization, we randomly select 128 samples with 2048 sequence length from the C4 dataset~\citep{C4}. We apply symmetric quantization to weights and asymmetric quantization to activations. Clipping thresholds for weight quantization are found via a  linear search over the squared error. Our scheme quantizes a 70B model in less than 2 hours on a single NVIDIA A100 GPU.

\subsection{Accuracy Recovery}

\paragraph{Accuracy Comparison on OPT.} We first compare the accuracy of QUIK with prior 4W4A quantization methods: SmoothQuant  \citep{xiao2022smoothquant}, RPTQ~\citep{yuan2023rptq} and OmniQuant~\citep{shao2023omniquant}.

\begin{table}[t]
    \centering
    \small
    \begin{tabular}{l|c|c|c|c}
        \multirow{2}{*}{Model} & \multicolumn{4}{c}{\textbf{OPT}} \\
        &  6.7B & 13B & 30B & 66B \\ \midrule
        Baseline  & 10.86& 10.13   &  9.56   &  9.34   \\ \midrule
        SmoothQuant & 1.8e4& 7.4e3 & 1.2e4 & 2.2e5 \\
        RPTQ  &17.83 & 17.83 &  11.50 & 11.16 \\
        OmniQuant   &12.24 & 11.65 & 10.60 & 10.29 \\
        QUIK (ours)  & \textbf{11.18} &  \textbf{10.78}  &  \textbf{10.08}   &  \textbf{9.66}  \\ 
    \end{tabular}
    \caption{Perplexity of 4-bit OPT models on the WikiText2 dataset. SmoothQuant, RPTQ, and OmniQuant results are taken from \citet{shao2023omniquant}, RPTQ denotes their improved numbers. Note that for the 66B model, all prior schemes keep  0.71\% of the linear layer operations in FP16 (the Head), while, by excluding outliers from quantization, we retain 2.78\% of operations in FP16.}
    \label{tab:accuracy_results_opt}
\end{table}

Table~\ref{tab:accuracy_results_opt} shows the results of all methods for 4 larger OPT models on the WikiText2 task~\citep{wikitext103}. We observed that, with QUIK, the accuracy of OPT models remains consistent even when employing a uniform number of outliers for all layers (instead of using a percentage of the input features). Consequently, we employed 256 outliers across all linear modules (which is $\approx 3\%$ of OPT-66B's hidden size). As can be seen, by effectively leveraging a small amount of full-precision outlier columns, QUIK can significantly outperform prior 4-bit methods, dropping only $0.3$ to $0.5$ points in perplexity relative to the full precision baseline. We emphasize that, for a fair comparison, QUIK quantizes \textit{all} linear backbone layers to 4-bit here. Additional results are presented in Appendix~\ref{appendix:full-accuracy-results-opt}.

\paragraph{Accuracy on LLaMA-2 and Falcon Models.} 
Next, we move to LLaMA-2 and Falcon models.
 See Table~\ref{tab:wikitext_results_LLaMA_falcon} for the results on WikiText2. As can be seen, QUIK-4B can preserve the accuracy in all models with at most $0.5$ perplexity loss for the LLaMA-2 models, and $0.3$ for Falcon models.

\begin{table}[h]
    \centering
    \resizebox{\columnwidth}{!}{%
    \begin{tabular}{l|c|c|c|c|c|c}
        \multirow{2}{*}{Model} & \multicolumn{3}{c|}{\textbf{LLaMA-2}}  & \multicolumn{3}{c}{\textbf{Falcon}}\\
        & 7B & 13B & 70B  & 7B & 40B & 180B \\ \midrule
        Baseline & 5.47 & 4.88 & 3.20 & 6.59 & 5.23 & 3.30\\ \midrule
        SmoothQuant  &  83.12  & 35.88  & -  & -  & -  & - \\  
        OmniQuant & 14.26  &  12.30 & -  & -  &  - & - \\  
        QUIK-4B & \textbf{5.84} & \textbf{5.28} & \textbf{3.74} & \textbf{6.90} & \textbf{5.46} & \textbf{3.61}\\
    \end{tabular}
    }
    \caption{Perplexity results of QUIK for 4-bit LLaMA-2 and Falcon models on WikiText2. We use 256 outliers for all linear layers. For the down-projection (in LLaMA-2 models) and FC2 layers (in Falcon models), we use 8-bit quantization, and increase the number of outliers (in FP16)  proportionally to the number of input features of these layers (which is not the case for other schemes). Results for SmoothQuant and OmniQuant follow~\citet{shao2023omniquant}. OmniQuant does not present results for the Falcon family and LLaMA2-70B in 4-bit. RPTQ does not present any results for LLaMA-2 and Falcon families.}
    \label{tab:wikitext_results_LLaMA_falcon}
\end{table}

\begin{table*}[h!]
    \centering
    \small
    \begin{tabular}{c|c|c|c|c|c|c|c}
        Model & Bits & Arc Challenge & Arc Easy & HellaSwag & PIQA & WinoGrande & Avg. Score \\
        
\midrule
\multirow{2}{*}{OPT-30B} & FP16 & 38.05 & 65.36 & 72.28 & 78.13 & 68.43& 64.45   \\
  & QUIK-4B &  36.69 & 64.39  & 70.84  & 77.75  & 67.01 &  63.34 \\

\midrule
\multirow{2}{*}{OPT-66B} & FP16 & 40.02  & 67.26  & 74.87  & 79.82  & 68.82 & 66.16 \\
  & QUIK-4B & 38.82  & 64.73  &  73.68 &79.43& 68.82 &  65.10  \\

\midrule
\multirow{2}{*}{LLaMA2-13B} & FP16 & 48.98  &  77.44 &79.38   & 80.52  & 72.22 & 71.70 \\
  & QUIK-4B  & 48.04  & 74.92  & 78.36  & 79.22  & 71.90 & 70.49  \\

\midrule
\multirow{2}{*}{LLaMA2-70B} & FP16 & 57.34  & 80.98  & 83.81  & 82.75  & 77.98  & 76.57 \\
  & QUIK-4B & 56.14  & 79.00  & 81.57  & 81.56  & 76.56 &  74.97 \\
    \end{tabular}
    \vspace{5pt}
    \caption{LM eval harness results of QUIK on OPT, LLaMA-2, and Falcon families. using 256 outliers.}
    \label{tab:zero_shots}
\end{table*}

\paragraph{Zero-Shot Accuracy.}
Next, we evaluate the impact of QUIK on the accuracy of zero-shot tasks. To this end, we study the 
average accuracy of the largest LLaMA-2 and OPT models on five popular zero-shot tasks: PIQA \citep{tata2003piqa}; WinoGrande \citep{sakaguchi2021winogrande}; HellaSwag \citep{zellers2019hellaswag}; Arc (Easy and Challenge) \citep{boratko2018systematic}. We use the LM Evaluation Harness \citep{gao2021framework} with default parameters in our experiments. Table~\ref{tab:zero_shots} shows the averaged accuracy of QUIK over zero-shot tasks. Similar to the generation task, QUIK preserves the accuracy of zero-shot tasks with at most a 1.5\% accuracy drop for LLaMA-2 models and 1.1\% for OPT models.

\paragraph{8-Bit Quantization.}
We compare the accuracy of QUIK-8B with SmoothQuant~\citep{xiao2022smoothquant} on OPT, LLaMA-2, and Falcon. We use asymmetric per-token quantization for activations and symmetric quantization for the weights in SmoothQuant (these are the same basic settings as for QUIK). Table~\ref{tab:int8_results} shows that although both schemes are close to lossless in terms of perplexity difference to FP16, QUIK produces higher accuracy results in most cases, compared to SmoothQuant. Further, it is unclear whether SmoothQuant can be applied to models with \emph{parallel attention}, such as the Falcon-7B model, where the MLP and Attention blocks share the same layer norm for their input, as this prevents scale factor fusion. See Appendix~\ref{appendix:full_int8_results} for further results.

\begin{table}[H]
    \centering
    \small
    \caption{Accuracy results for 8-bit models on WikiText2. We use 256 outliers in QUIK experiments. Following the SmoothQuant paper, we use $\alpha=0.8$ for LLaMA-2 models and $\alpha=0.5$ for OPT and Falcon families.}
    \vspace{5pt}
    \begin{tabular}{l|c|c|c|c|c|c}
        \multirow{2}{*}{Model} &   \multicolumn{2}{c|}{\textbf{OPT}}  & \multicolumn{2}{c|}{\textbf{LLaMA-2}}  & \multicolumn{2}{c}{\textbf{Falcon}}\\
        &  30B & 66B  &  13B & 70B &  40B& 180B \\ \midrule
        FP16 &   9.56 & 9.34 &  4.88 & 3.20 & 5.23 & 3.30  \\  \midrule
        SmoothQuant &  9.59 & 9.80 & 4.94 & 3.48 & 5.26 & \textbf{3.30} \\  \midrule
         QUIK-8B &    \textbf{9.51} & \textbf{9.29} &  \textbf{4.89} & \textbf{3.33} & \textbf{5.23} & 3.31\\  
    \end{tabular}
    \label{tab:int8_results}
\end{table}
\setlength\tabcolsep{5pt}

\paragraph{Outlier-Free Layers.}
Finally, we study the effect of keeping multiple linear layers without any outliers.
This might help boost end-to-end performance by removing all
the outlier-related overheads during the forward pass. (Although, as we show later, these overheads are minor.) Table~\ref{tab:wikitext2_thresholding} shows how the accuracy of different models changes when we use different absolute threshold values (shown by \textbf{T}), extracted using a linear search, for the outliers. We conclude that there is no universal threshold across all models, which would preserve accuracy across all models. For example, Falcon-180B can achieve reasonable accuracy even if 24\% of the linear layers (115 out of 480) contain zero outliers. However, this is not the case for smaller models: LLaMA2-70B can recover accuracy with up to 5\% of the linear layers (30 out of 560) having zero QUIK outliers. We provide additional experiments in Appendix~\ref{appendix:zero-outlier_full}.

\begin{table}[]
    \centering
    \small
    \begin{tabular}{l|c|c|c}
        \multirow{1}{*}{Model} & \multirow{1}{*}{\textbf{T}} & LLaMA2-70B & Falcon-180B \\
        \midrule
        FP16 & - &   3.2  & 3.30  \\ \midrule
        \multirow{7}{*}{QUIK-4B} & 0 &   3.74 (0) & 3.61 (0) \\
        \cmidrule{2-4}
        & 2.0 &   3.75 (10) &  3.61 (3)  \\
        \cmidrule{2-4}
        & 3.0 &  3.85 (30) &  3.61 (4)\\
        \cmidrule{2-4}
        & 4.0 &    5.15 (58) &   3.72 (14)\\
        \cmidrule{2-4}
        & 8.0 &  5.92 (219) & 3.73 (115)\\
    \end{tabular}
    \vspace{5pt}
    \caption{
    Study of zero outlier setting on WikiText2 using 256 outliers. We use zero outliers when the maximum of scale is less than threshold \textbf{T}. For each experiment, the number of linear layers with zero outliers is written in parentheses.}
    \label{tab:wikitext2_thresholding}
    \vspace{-5pt}
\end{table}

\vspace{-0.5em}
\subsection{Performance Analysis}
\label{sec:speedup_results}
We now examine the performance of the QUIK implementation by evaluating different aspects of our kernel. We use PyTorch/1.13, CUDA/11.8, Huggingface Transformers/4.34. We run all our experiments on RTX 3090 GPUs as our main goal is to accelerate LLM inference on commodity GPUs. Appendix~\ref{appendix:rtx3080} shows similar results on RTX 3080 GPUs. 

\paragraph{Peak Memory Usage.} First, we assess the memory usage of our quantized models. 
In Table \ref{tab:memory_usage}, we evaluate the peak memory usage across different configurations for the OPT and LLaMA-2 families. For OPT-66B, the QUIK-8B and QUIK-4B models demonstrate peak memory reductions of approximately 47\% (compared to the ideal 50\% reduction) and 74\% (compared to the ideal 75\% reduction), respectively. For the LLaMA2-70B model, the reductions are 32\% for QUIK-8B and 67\% for QUIK-4B. This is because we keep the down-projection in 8-bits and use additional outliers. Additional overheads come from auxiliary buffers, which differ for various layer sizes.

\begin{table}[]
    \centering
    \begin{tabular}{l|c|c|c|c|c|c}
        \multirow{2}{*}{Model} & \multicolumn{3}{c|}{\textbf{OPT}}  & \multicolumn{3}{c}{\textbf{LLaMA-2}} \\
        & 13B & 30B & 66B & 7B & 13B & 70B \\ \midrule
        Baseline               & 30.5 & 67.4 & 162.1 & 14.9 & 28.0 & 147.1 \\ \midrule
        QUIK-8B   & 16.1 & 39.3 & 81.2 & 14.6 & 25.2 & 99.3 \\ \midrule
        QUIK-4B & \textbf{10.7} & \textbf{24.6} & \textbf{45.1} & \textbf{7.1} & \textbf{12.1} &  \textbf{49.1} \\
    \end{tabular}
    \caption{Peak memory usage (in GB) in an end-to-end benchmark. In total, the outliers take 2.71 GB and 4.06 GB for OPT-66B and LLaMA2-70B models respectively.}
    \label{tab:memory_usage}
\end{table}

\paragraph{Ideal and Layer-wise Speedups.} 
Next, we evaluate the ideal speedups, as well as the actual speedups we measure in each Transformer block separately. The results in Figure~\ref{fig:ideal-speedups} depict ``ideal'' computational power for layer-wise matrix multiplications at different precision levels, without taking into account any quantization/dequantization overheads. 
Here, we focus on realizable speedups when executing Algorithm~\ref{alg:kernels}, which includes mixed-precision multiplication as well as compression and decompression operations.

\begin{figure}[t!]
\centering
\includegraphics[width=0.35\textwidth]{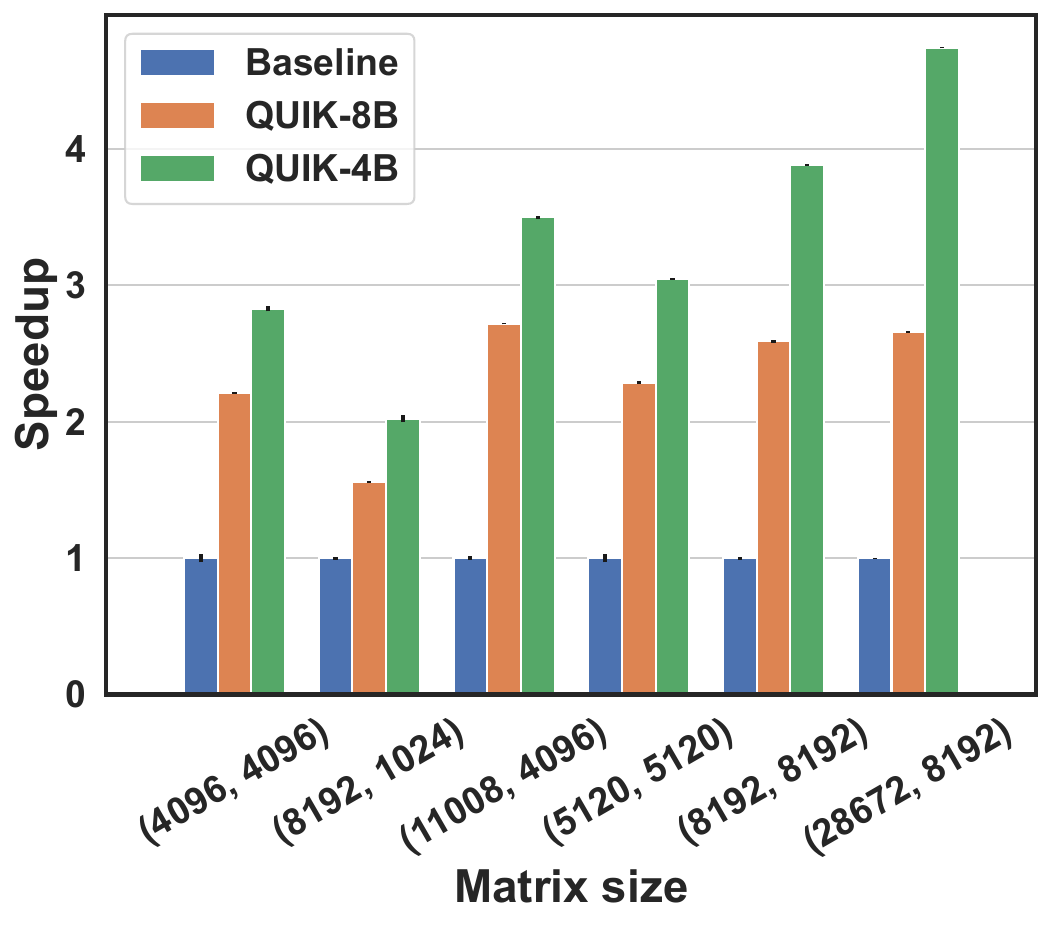}
\caption{\label{fig:bench_layerwise} Layer-wise speedups on a single RTX3090 for different layer sizes and compression types. QUIK-4B with 256 outliers, QUIK-8B without outliers.}
\vspace{-.5em}
\end{figure}

In Figure~\ref{fig:bench_layerwise}, we compare the layer-wise performance of quantized linear layers (QUIK-4B uses 256 outliers per layer) relative to FP16, for a full implementation of our algorithm. The matrix sizes correspond to layers in LLaMA models. We observe that QUIK-4B can achieve slightly higher than $4\times$ speedup on large layers and over $2\times$ on smaller ones. Thus, the speedups of raw low-precision matmul speedups can partially ``hide'' the overheads of QUIK.

\paragraph{End-to-end speedups.} Finally, we also demonstrate the end-to-end speedup benefits of QUIK models. For this purpose, we integrate QUIK into the widely used HuggingFace PyTorch implementation, by replacing linear layers with 4-bit (and 8-bit) QUIK re-implementations. For the LLaMA model, we use  FlashAttention~\citep{dao2022flashattention} for all models (including  FP16). The number of outliers in QUIK-4B is set to 256 except for the special case of down projection layers in LLaMA and FC2 in the Falcon models, which we quantize to 8 bits with more than 600 outliers.

\begin{figure}[t]
    \centering
    \setkeys{Gin}{width=0.49\linewidth}
    \begin{subfigure}
        \centering
        \includegraphics{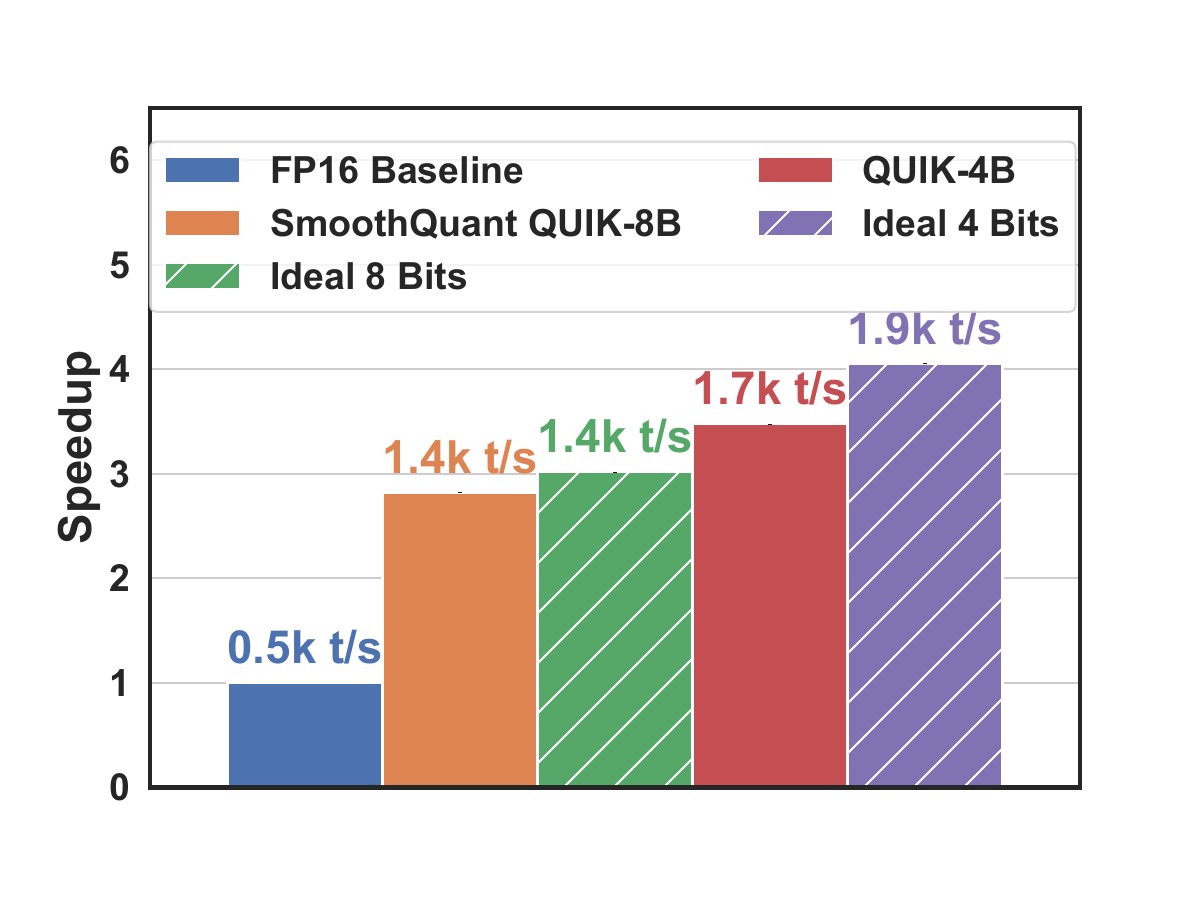}
    \end{subfigure}\hfil
    \begin{subfigure}
        \centering
        \includegraphics{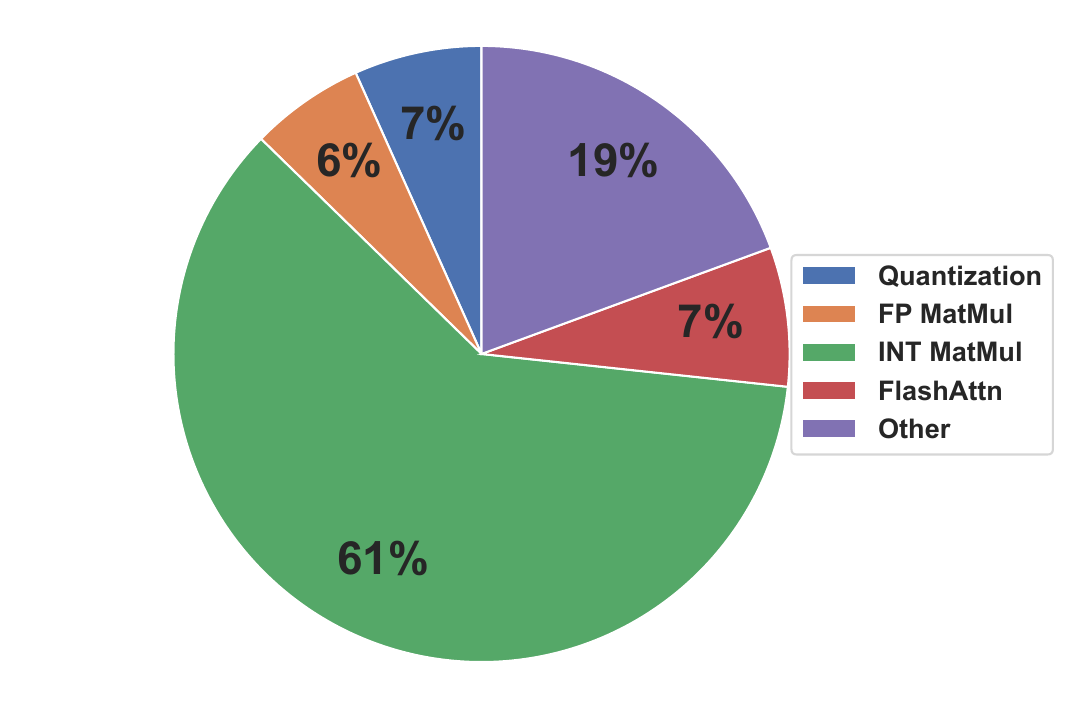}
    \end{subfigure}
    \caption{Performance results and overhead breakdown on LLaMA2-70B on a machine with 8x RTX 3090 GPUs. \textbf{Left:} Speedup vs. FP16 and vs. an ideal implementation, without overheads, for 4-bit and 8-bit QUIK kernels with absolute throughput values. \textbf{Right:} Performance breakdown of end-to-end inference benchmark for QUIK-4B with outliers in terms of MatMul time vs. quantization overheads.}
    \label{fig:LLaMA-70B}
\end{figure}

\begin{figure*}[h]
    \centering
    \setkeys{Gin}{width=0.33\linewidth}
    \subfigure[OPT \label{fig:opt_end2end}]{\includegraphics{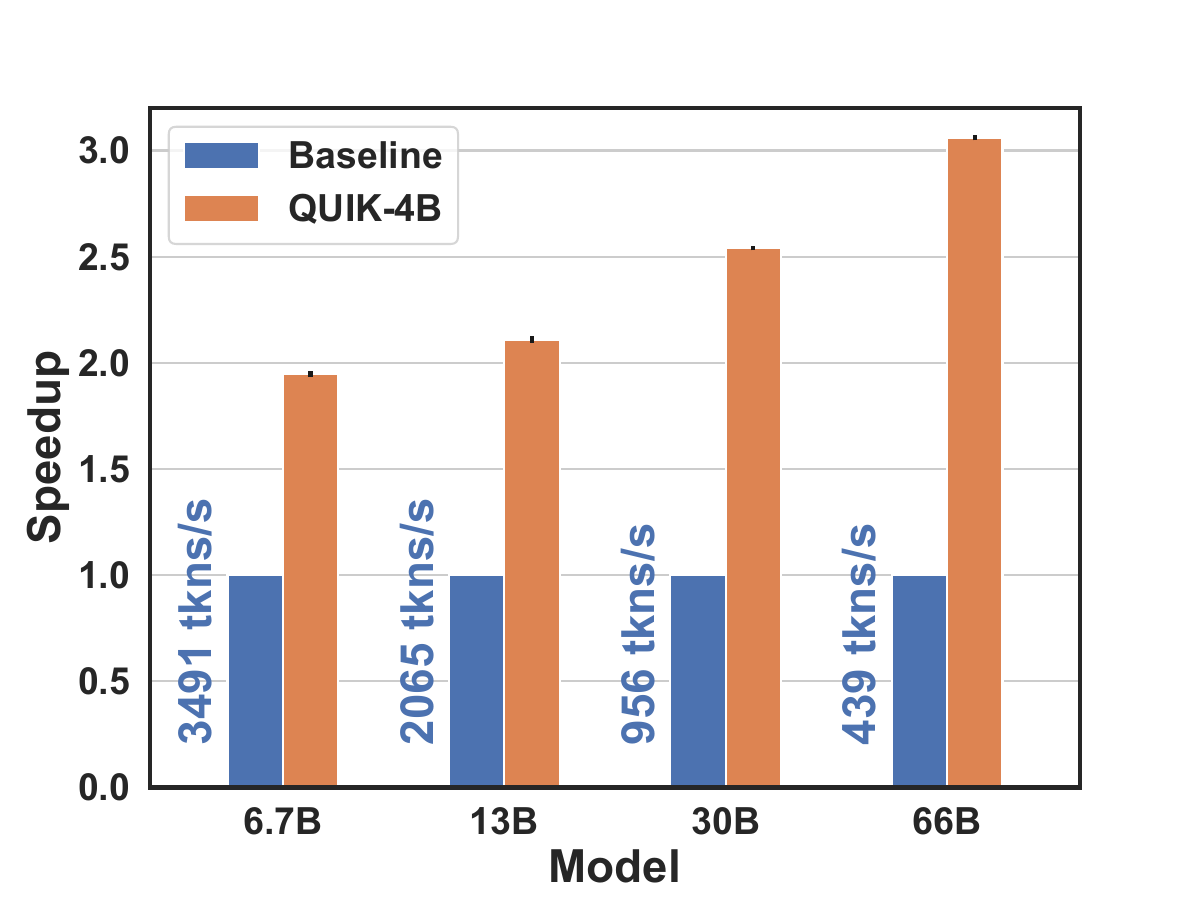}}\hfil
    \subfigure[LLaMA-2 \label{fig:LLaMA_end2end}]{\includegraphics{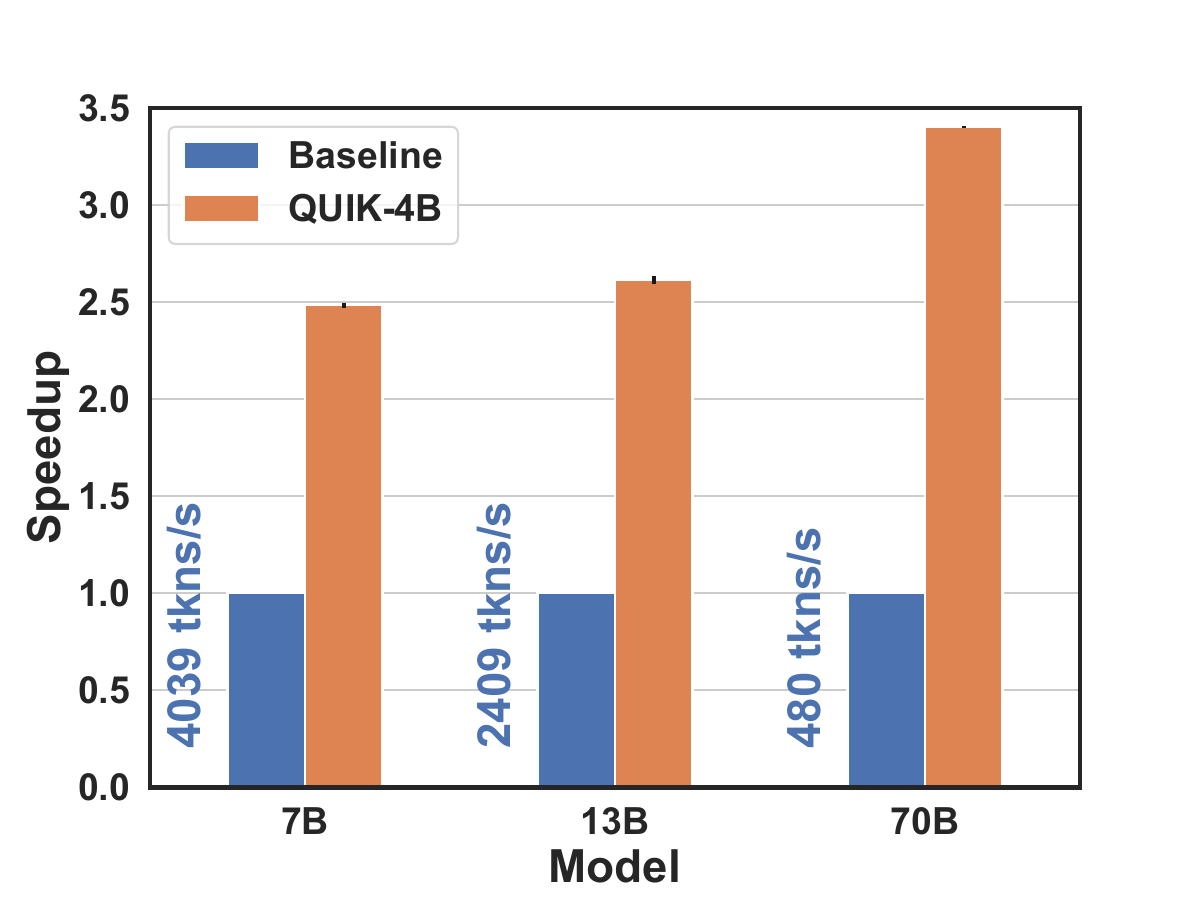}}\hfil
    \subfigure[Falcon \label{fig:falcon_end2end}]{\includegraphics{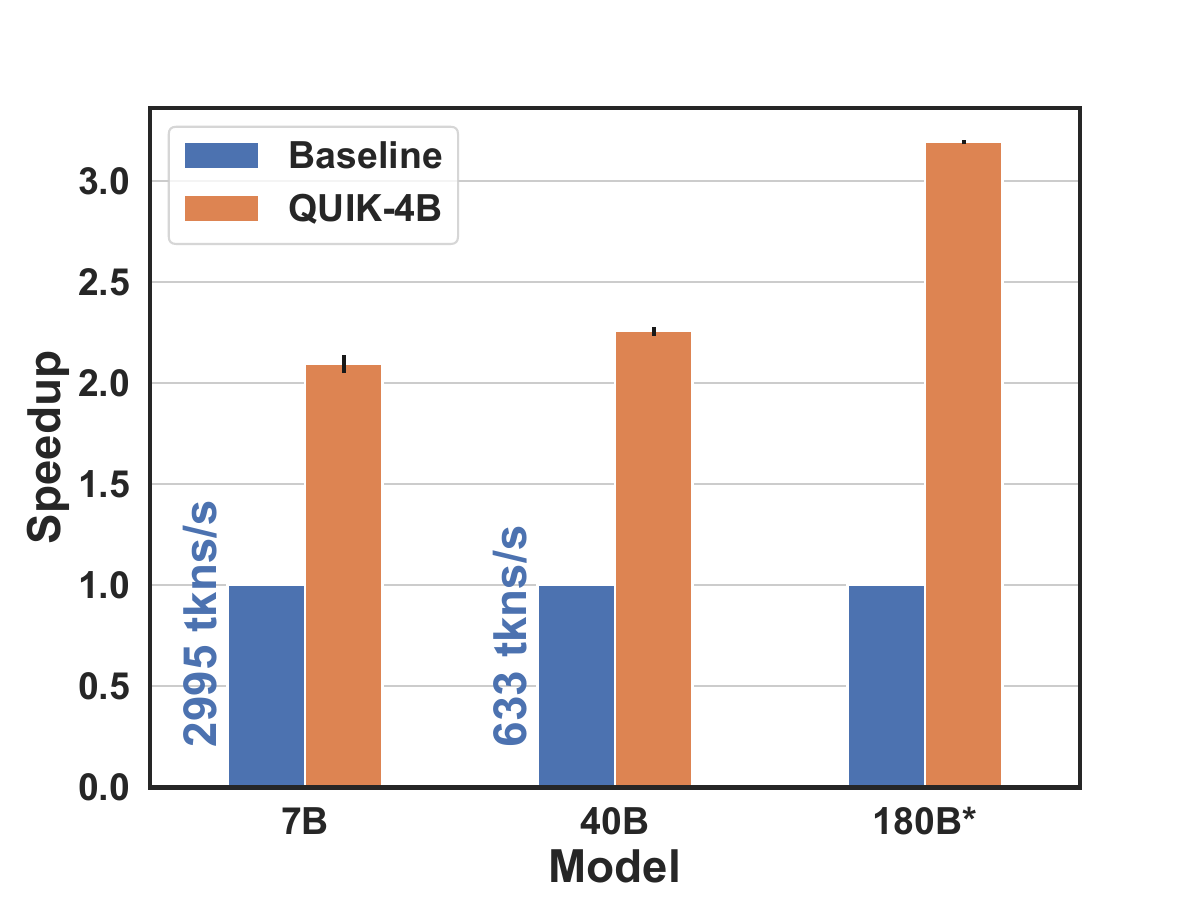}}
    \caption{End-to-end inference speedups for QUIK-4B with outliers relative to the FP16 baseline, on NVIDIA RTX 3090 GPUs. Falcon-180B results are from single Transformer block inference benchmark.}
    \label{fig:perf_end2end}
\end{figure*}

In Figure ~\ref{fig:perf_end2end}, we compare the throughput improvements of prefill passes (for single batches with 2048 tokens) for quantized models, relative to the corresponding FP16 version. The bar plot shows throughput improvements of QUIK-4B compared to FP16. The annotations to the baseline represent its actual throughput values in our experiments. For instance, OPT-66B using FP16 linear layers achieved 439 tokens/s whereas the same model inference with QUIK-4B linear layers resulted in 1343 tokens/s. This shows that, in addition to a close to $4\times$ memory reduction, which reduces the number of required GPUs for inference, QUIK also achieves up to $3.4\times$ higher throughput relative to FP16, with the biggest improvements attained on the largest models (LLaMA2-70B), where the relative impact of overheads is lowest. The memory reduction is important in the Falcon inference case: we were not able to run Falcon-180B in full precision on 8xRTX3090 GPUs, as the max memory peak of the model is more than 360GB. However, QUIK-4B allows us to run full inference of this 180B model on a single server resulting in 542 tokens/second. Therefore, we estimated speedups for the FP16 180B model in Figure~\ref{fig:falcon_end2end} based on the runtime of a single Transformer block.

We emphasize that the speedups in our end-to-end experiments are exclusively through QUIK accelerated linear layers. All other functions are precisely the same. 
As shown in Figure~\ref{fig:LLaMA-70B} (Right), the overheads from attention, softmax, or layernorm operations become significant when a large fraction of the computation occurs in 4-bit precision.

\begin{figure}[]
\centering     
\includegraphics[width=\linewidth]{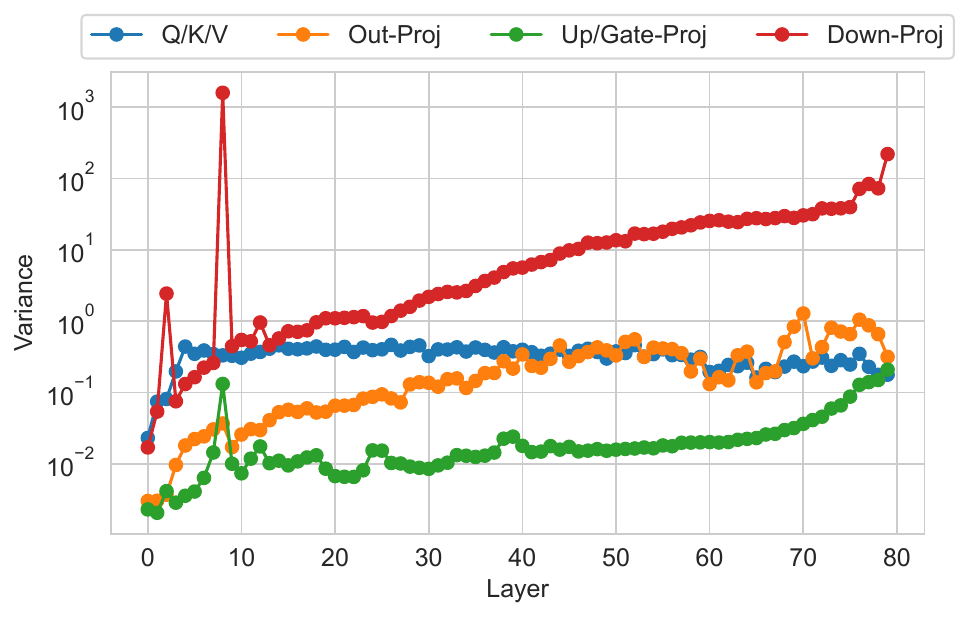}
\caption{
The variance of the inputs in different layers of LLaMA2-70B. The "Down-Proj" layers have significantly larger variances, resulting in poor 4-bit quantization. 
}
\label{fig:LLaMA70b_var_max}
\vspace{-1mm}
\end{figure}

\begin{figure}[t]
\centering
\includegraphics[width=0.9\columnwidth]{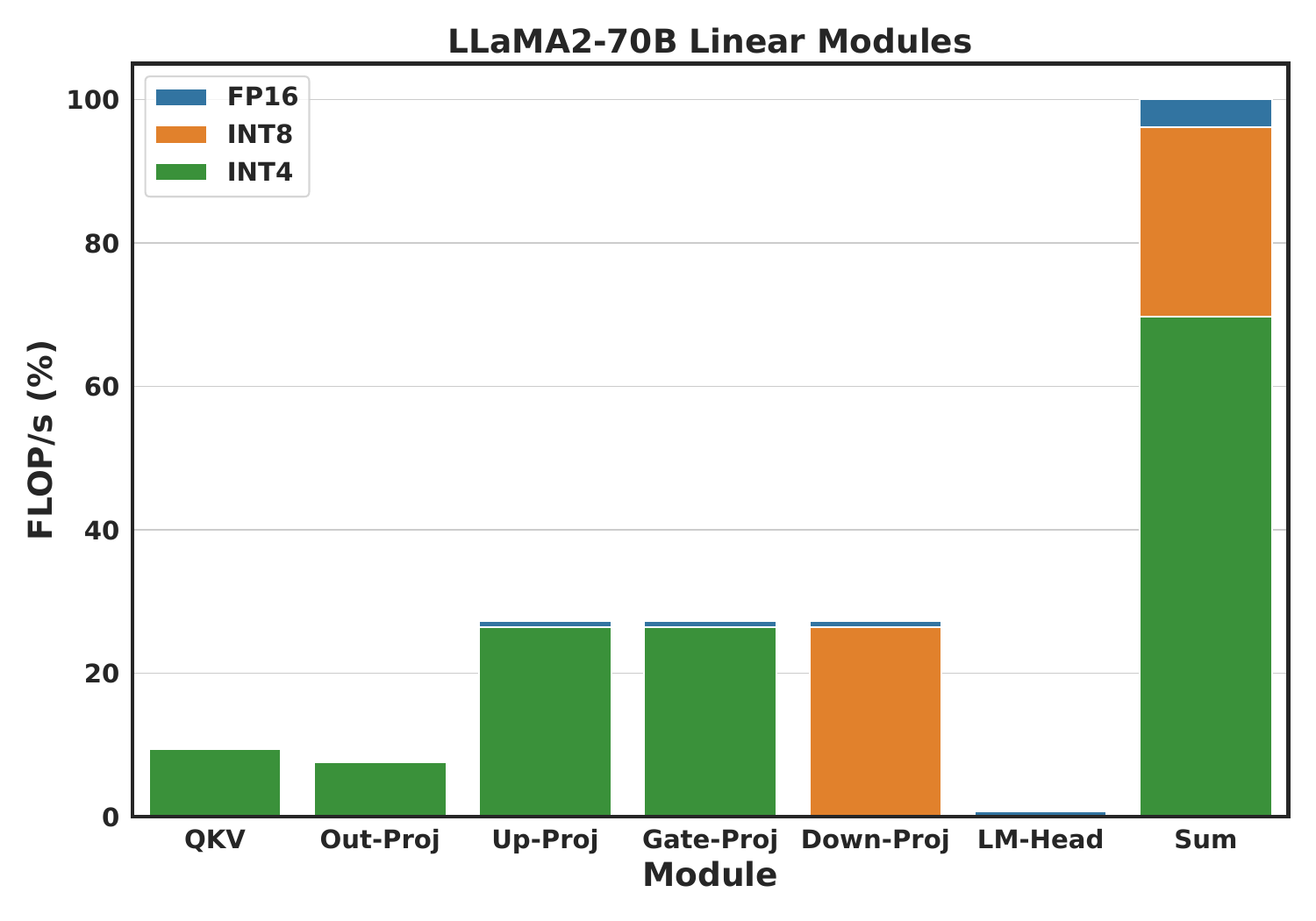}
\caption{
FLOP/s analysis of the LLaMA2-70B linear layers with QUIK. We use 3.125\% outliers (256 outliers in all layers and 896 for the down-projection layer) and 2048 sequence length.}
\label{fig:llama70b_flops}
\end{figure}

\paragraph{Outlier Performance Costs.} To illustrate the performance implications of supporting outliers, in Figure~\ref{fig:LLaMA-70B} (left) we provide end-to-end speedups for variants of the HuggingFace integration where we directly use 8-bit and 4-bit kernels, without preserving accuracy (Ideal 8-bit and 4-bit), relative to our accuracy-preserving QUIK implementations. 

We observe that the 8-bit implementation provides close to ideal speedups, reducing the number of GPUs from 7 to 5. 
QUIK-4B (taking outliers into account) performs $\approx$15\% better, further reducing the number of required GPUs to 3, using less than 50 GB of GPU memory. 
The performance impact of outlier selection (hence mixed precision matrix multiplication) and selective 8-bit quantization (for down-projection MLP layer) is shown in the comparison with Ideal 4-bit. 
QUIK-4B is within 15\% of Ideal 4-bit performance. (However, it is currently not known how a model with weights and activations in 4 bits could recover accuracy.) 
The justification for this performance impact is provided in Figure~\ref{fig:LLaMA-70B} (right), where we break down the per-operation overheads for LLaMA2-70B inference. Specifically, we observe here and in Figure~\ref{fig:optim_profile} that the overheads of quantization and full precision multiplication can take up a large fraction of the overall operation time, especially for smaller matrices. 
This illustrates the trade-offs between performance and accuracy for a specific model.

\subsection{Ablation Studies}
\label{sec:ablation}
We now provide in-depth examples for using QUIK on two large models: LLaMA2-70B, and Falcon-180B. The former model is important as it shows high performance across different tasks~\citep{touvron2023llama2}. The latter is the largest openly-available GPT-type model.

\subsubsection{Case Study 1: LLaMA2-70B}
\label{sec:llama-case-study}

First, we study the FLOP breakdown across precisions using QUIK-4B on LLaMA2-70B. Next, we study the effect of key parameters of QUIK: 8-bit Down-Projection, and Outlier Counts. We provide additional ablation in Appendix~\ref{appendix:full-accuracy-results-LLaMA}.

\paragraph{8-bit Down-Projection.}
Within the MLP module of the LLaMA2-70B model, three linear layers are present, referred to as "Up-Proj", "Gate-Proj", and "Down-Proj".  "Up-Proj" and "Gate-Proj" share an input (MLP input) and apply their respective linear transformations to it. Subsequently, the output of "Gate-Proj" is subjected to a SiLU activation function. Lastly, the input for the "Down-Proj" layer is constructed by taking the Hadamard product of the outputs from "Up-Proj" and "Gate-Proj".

\begin{table}[H]
    \centering
    \small
    \begin{tabular}{l|c|c|c}
        \textbf{LLaMA-2} &  7B & 13B & 70B \\ \midrule
        Baseline & 5.47 & 4.88 & 3.20 \\ \midrule
        QUIK-4B & 5.84 & 5.28 & 3.74\\
        4-bit Down-Proj & 8.87 & 7.78 & 6.91   \\
    \end{tabular}
    \caption{Ablation for keeping the down-projection layer in 4-bits.}
    \label{tab:LLaMA_70b_down_proj_ablation}
\end{table}

Figure \ref{fig:LLaMA70b_var_max} shows the variance of the input across various layers in LLaMA2-70B, which we use as a guide to choose both the number of outliers and the set of layers to be executed in 8 bit precision. 
Specifically, it can be observed that the "Down-Proj" layers have large input variance,
mainly due to the Hadamard product of the previous two outputs, resulting in poor accuracy for 4-bit quantization. To address this, we employ \emph{8-bit quantization} for both the weights and activations within the "Down-Proj" layers of LLaMA2 models. Table~\ref{tab:LLaMA_70b_down_proj_ablation} shows that keeping the down-projection
layers in 8-bit is critical for high accuracy on
LLaMA2, as it improves perplexity by $>$ 2 points, across all models.

\paragraph{FLOP/s Analysis.} 
Figure~\ref{fig:llama70b_flops} shows the percentage of the FLOP/s we keep in each precision (INT4 for base weights, FP16 for outliers, and INT8 for down-projection layers) in LLaMA2-70B. More precisely, for 256 outliers, we perform $\approx$70\% of the operations in 4-bit and $\approx$27\% using 8-bits.

\begin{table}[H]
    \centering
    \small 
    \begin{tabular}{l|c|c|c}
        \multirow{2}{*}{Method}  & \multirow{2}{*}{Outliers}  & Down-Proj &WikiText2 \\ 
         &  & Outliers & (PPL)\\ \midrule
        Baseline & - & - &3.20 \\ \midrule
        \multirow{4}{*}{QUIK-4B} & 128 & 448 & 3.80 \\
        & 256 &896 & 3.74\\
        & 512 & 1792& 3.67 \\
        & 1024 &3584 & 3.62 \\
    \end{tabular}
    \caption{Ablation study of different outlier numbers in QUIK for the LLaMA2-70B model.}
    \label{tab:LLaMA70b_outlier_num}
\end{table}

\paragraph{Outlier Count.}
\label{sec:outlier_count}
Finally, we look at how different outlier counts affect the WikiText2 score for the LLaMA2-70B model. In Table~\ref{tab:LLaMA70b_outlier_num}, we observe that increasing the outliers from 128 to 1024 results in a 0.2 perplexity improvement. 
We also adjusted the outliers for down-projection layers, ensuring there are 3.5x times more than the other linear layers, to match input size. Our results show that using 256 outliers is already a good  choice for our experiments. Using additional outliers does not significantly improve accuracy.

\subsubsection{Case Study 2: Falcon-180B}
\label{sec:falcon-case-study}
In this section, we revisit  applying QUIK to Falcon-180B, the largest GPT-style openly-available model.
The model requires $\approx 365$GB of GPU memory for the inference, which makes it impossible to run inference on a GPU server with 8x RTX3090 nodes (192 GB memory), illustrating the importance of reducing the memory footprint of this model. 

The results in Tables~\ref{tab:wikitext_results_LLaMA_falcon} and~\ref{tab:wikitext2_thresholding}, and Figure~\ref{fig:perf_end2end} already presented accuracy and performance results for this model for QUIK variants. 
Here, we investigate leveraging the hardware-supported 2:4 sparse + INT4 format by combining QUIK with 2:4 sparsity  for this model.

\paragraph{Joint INT-4 Quantization and 2:4 Sparsification.} 
A simple solution for pushing the limits of the model compression is to sparsify the already quantized model (or vice-versa). However, this results in high accuracy drops. Instead, we extend the SparseGPT algorithm~\citep{frantar2023sparsegpt}  
to support our outlier scheme to jointly quantize and sparsify the model, while keeping the outlier features in dense FP16. In Table~\ref{tab:sparsity_quant_falcon180b}, we present the results of quantizing all layers, but selectively keep certain layer types dense. Specifically, we found that one-shot pruning of the weights in the attention blocks to the 2:4 pattern throughout all layers largely preserves accuracy, leading to small memory gains. We present 8-bit results in the same setting in  Appendix~\ref{appendix:int8_sparse_full_results}.

\begin{table}[]
    \centering
    \resizebox{\columnwidth}{!}{%
    \begin{tabular}{l|c|c|c|c}
        \multirow{2}{*}{Precision}  & \multirow{2}{*}{Sparsity}  & Dense   &  WikiText2  & Mem. Peak  \\ 
        & & Layers& (PPL) & (rel to FP16)\\ \midrule
        \multirow{2}{*}{FP16} & 0\% & All & 3.30 & 100\%\\ 
         & 2:4 & None & 6.13 & - \\ \midrule
        \multirow{4}{*}{QUIK-4B} & 0\% & All & 3.61 & 38 \%\\ 
         & 2:4 & None & 6.62 & 25\% \\ 
         & 2:4 & Attn. Blocks & 6.34 & 26\% \\ 
         & 2:4 & MLP Blocks &  \textbf{3.93} & 36\% \\ 
    \end{tabular}
    }
    \caption{Accuracy results for quantized + 2:4 sparsified on Falcon-180B. For the quantized experiments, we apply quantization on all layers with 256 outliers but keep some of the layers in dense (mentioned in the Table). By memory peak we mean the maximal amount of allocated memory (in GB) during the inference of a single Transformer block.}
    \label{tab:sparsity_quant_falcon180b}
\end{table}

%% file: conclusion.tex
We presented a hybrid quantization scheme called QUIK, 
executing a large majority of inference computation in 4-bit precision, with efficient GPU support. We have shown significant speedups using QUIK across several LLM types, on commodity hardware. In future work, we plan to examine a unified implementation which would support both single-token and multi-token inference on top of QUIK weights, integration with speculative decoding~\cite{leviathan2023fast}, and additional models.

%% file: appendix.tex
\onecolumn

\section{Full OPT Accuracy Results}
\label{appendix:full-accuracy-results-opt}

Table \ref{tab:opt_full_ppl} shows the perplexity results of OPT models. We use symmetric quantization for the weights in all our experiments. The results suggest that in a 4-bit setting, 
considering outlier features is crucial to preserve the accuracy even in small models (like OPT-1.3b). We note that 256 outliers is equivalent to 12.5\% of the 1.3B model's hidden size (and 2.77\% of the 66B model's hidden size).

\begin{table*}[h]
\centering
\vspace{2mm}
\resizebox{\linewidth}{!}{
\begin{tabular}{l|ccc|ccc|ccc|ccc|ccc}
Model  & \multicolumn{3}{c|}{OPT-1.3b} & \multicolumn{3}{c|}{OPT-6.7b} & \multicolumn{3}{c|}{OPT-13b} & \multicolumn{3}{c|}{OPT-30b} & \multicolumn{3}{c}{OPT-66b} \\ \midrule
Task &   WIKI     & PT      & C4      & WIKI     & PT      & C4      & WIKI    & PT      & C4      & WIKI    & PT      & C4      & WIKI    & PT      & C4      \\ \midrule
Baseline  &  14.63    & 16.96   & 14.72  &   10.86 & 13.09   &  11.74  &  10.13 & 12.34    & 11.20   &  9.56  & 11.84   & 10.69   &  9.34  & 11.36   &  10.28  \\ \midrule

GPTQ-4B     &  \multirow{1}{*}{15.89}   &  \multirow{1}{*}{18.83}   &  \multirow{1}{*}{15.90} & 
\multirow{1}{*}{11.43}  &  \multirow{1}{*}{13.81}   &  \multirow{1}{*}{12.21}  & 
\multirow{1}{*}{10.38}  &  \multirow{1}{*}{12.65}    &  \multirow{1}{*}{11.41}   & 
\multirow{1}{*}{9.60}  & \multirow{1}{*}{12.02}    & \multirow{1}{*}{10.83}   &
\multirow{1}{*}{9.65}  &  \multirow{1}{*}{11.63}   &  \multirow{1}{*}{10.56} 
 \\   \midrule

0 Outliers &   15k   &  9k  & 10k  &  10k &  9k  & 9k   & 9k  & 12k    & 9k   &  12k  & 13k   & 17k   & 12k  & 13k   &  10k  \\ 
64 Outliers &  26.259 & 27.143 & 22.981  & 11.473 & 13.888 & 12.348   & 11.031 & 13.305 & 11.971   & 10.283 & 12.557 & 11.267  &  9.851 & 11.965 & 10.742  \\ 
128 Outliers &  17.638 & 19.709 & 16.799    & 11.671 & 13.809 & 12.314  & 10.964 & 13.241 & 11.894   &  10.339 & 12.564 & 11.279    &  9.805 & 11.842 & 10.653   \\ 
256 Outliers &   17.358 & 19.525 & 16.607  &  11.184 & 13.811 & 12.262  & 10.779 & 13.175 & 11.847   &  10.078 & 12.465 & 11.226    &  9.662 & 11.793 & 10.635   \\

\end{tabular}
}
\caption{Perplexity scores of QUIK-4B over various OPT models with different outliers on three datasets: WikiText2 (WIKI), Pen Treebank (PT), and C4. GPTQ-4B only quantizes the weights (using int-4 symmetric quantization) and keeps the activations in FP16.}\label{tab:opt_full_ppl}
\vspace{5mm}
\end{table*}

\section{Full LLaMA-2 Accuracy Results}
\label{appendix:full-accuracy-results-LLaMA}

Table \ref{tab:LLaMA-wikitext2} shows the perplexity of QUIK on LLaMA-2 models. 
We provide a list of tricks to improve the quality of the model without too 
much overhead. We found that keeping the down-proj layer in 8 bits can 
improve the perplexity by about 3 points. Also, we found weight clipping
as a cheap and efficient trick for improving the accuracy of QUIK-4B.

\begin{table}[H]
    \centering
    \begin{tabular}{l|c|c|c|c|c}
        LLaMA-2 & Down-Proj & Clipping & 7B & 13B & 70B \\
        \midrule
        FP16 & W16A16 & - & 5.47 & 4.88 & 3.2 \\
        \midrule
        GPTQ-4B & W4A16 & - & 6.24 & 5.25 & 3.68  \\
        \midrule
        QUIK-4B  & W4A4 & - & 8.78 & 7.78 & 6.91  \\
        \midrule
        QUIK-4B  & W4A16 & - & 6.09 & 5.49 &   3.98 \\
        \midrule
        QUIK-4B & W4A8 & - & 6.11  &  5.5  &  4.0  \\
        \midrule
        QUIK-4B  & W8A8 & - & 5.98  &  5.37  & 3.87  \\
        \midrule
        QUIK-4B & W8A8 &\checkmark & 5.84   & 5.28    &  3.74  \\
    \end{tabular}
    \vspace{5pt}
    \caption{
    LLaMA-2 perplexity results on WikiText2 using 256 outliers. We apply clipping only during the weight quantization.} \vspace{5mm}
    \label{tab:LLaMA-wikitext2}
\end{table}

\section{Full INT-8 Accuracy Results}
\label{appendix:full_int8_results}

Table~\ref{tab:int8_results_full} shows QUIK-8B comparison against SmoothQuant on the WikiText2 dataset. We use per-token (per-column) quantization for the activations (weights) in SmoothQuant and only apply the quantization on the linear layers (which is the case for QUIK also). We exclude the Falcon-7B model as this model has a single layer-norm for both MLP and Attention blocks and it is not clear how the weights of the FC1 and KQV will be updated in the SmoothQuant algorithm.

\begin{table*}[h]
    \centering
    \small
    \vspace{5pt}
    \begin{tabular}{l|c|c|c|c|c|c|c|c|c|c}
        \multirow{2}{*}{Model} &   \multicolumn{5}{c|}{\textbf{OPT}}  & \multicolumn{3}{c|}{\textbf{LLaMA-2}}  & \multicolumn{2}{c}{\textbf{Falcon}}\\
        & 1.3b & 6.7B & 13B & 30B & 66B  & 7B & 13B & 70B &  40B& 180B \\ \midrule
        FP16 &  14.63 & 10.84 & 10.13 & 9.56 & 9.34 & 5.47 & 4.88 & 3.20 & 5.23 & 3.30  \\  \midrule
        SmoothQuant &  14.70 & 10.89 & 10.37 & 9.59 & 9.80 & 5.58 & 4.94 & 3.48 & 5.26 & \textbf{3.30} \\  \midrule
         QUIK-8B &  \textbf{14.62} & \textbf{10.84} & \textbf{10.13} &  \textbf{9.51} & \textbf{9.29} & \textbf{5.48}  &  \textbf{4.89} & \textbf{3.33} & \textbf{5.23} & 3.31\\  
    \end{tabular}
    \caption{Accuracy results for 8bit models on WikiText2. We use 256 outliers in QUIK experiments. Following the SmoothQuant paper, we use $\alpha=0.8$ hyperparameter for LLaMA-2 models and $\alpha=0.5$ for OPT and Falcon families.}
    \vspace{5mm}
    \label{tab:int8_results_full}
\end{table*}
\setlength\tabcolsep{5pt}

\section{Zero-Outlier Full Results}
\label{appendix:zero-outlier_full}
Table~\ref{tab:wikitext2_thresholding_full} shows the results of keeping different numbers of layers without outliers for different models.

\begin{table*}[]
    \centering
    \begin{tabular}{c|c|c|c|c|c|c|c}
        \multirow{2}{*}{Model} & \multirow{2}{*}{\textbf{T}} & \multicolumn{3}{c|}{\textbf{LLaMA-2}} & \multicolumn{3}{c}{\textbf{Falcon}} \\
        & &7B &13B & 70B& 7B & 40B& 180B \\
        \midrule
        FP16 & - &  5.47 & 4.88 & 3.2 & 6.59 & 5.23 & 3.30  \\
        \midrule
        \multirow{7}{*}{QUIK-4B} & 0 &   5.84 (0)  &  5.28 (0)  & 3.74 (0) & 6.90 (0)& 5.46 (0) & 3.61 (0) \\
        \cmidrule{2-8}
        & 2.0 &  5.91 (5) &  5.33 (3)  & 3.75 (10) & 6.90 (3) & 5.46 (1) & 3.61 (3)  \\
        \cmidrule{2-8}
        & 3.0 & 6.09 (11) &  5.34 (8)  & 3.85 (30) & 6.91 (14) & 5.46 (2) & 3.61 (4)\\
        \cmidrule{2-8}
        & 4.0 &  6.13 (21) & 5.36 (17)   & 5.15 (58) & 6.93 (27) & 10.56 (8) & 3.72 (14)\\
        \cmidrule{2-8}
        & 8.0 &  12.93 (55) &  21.85 (66)  & 5.92 (219) & 6.94 (57) & 10.61 (33) & 3.73 (115)\\
    \end{tabular}
    \vspace{5pt}
    \caption{
    Study of zero outlier setting on WikiText2 using 256 outliers. We use zero outliers when the maximum of scale is less than threshold \textbf{T}. For each experiment, the number of linear layers with zero outliers is written in parentheses.}
    \label{tab:wikitext2_thresholding_full}
\end{table*}

\section{2:4 Sparsity + INT8 Quantization}
\label{appendix:int8_sparse_full_results}
Table~\ref{tab:sparse_int8quantized_results_full} shows the accuracy results of applying QUIK-8B with 2:4 sparsity across all models. The results suggest that the main accuracy drop is from introducing 2:4 sparsity to the weight matrices and keeping some of the layers in dense is crucial to preserve the accuracy (See section~\ref{sec:falcon-case-study}).

\begin{table*}[h]
    \centering
    \small
    \vspace{5pt}
    \begin{tabular}{l|c|c|c|c|c|c|c|c|c|c|c|c}
        \multirow{2}{*}{Model} & \multirow{2}{*}{Sparsity} &   \multicolumn{5}{c|}{\textbf{OPT}}  & \multicolumn{3}{c|}{\textbf{LLaMA-2}}  & \multicolumn{3}{c}{\textbf{Falcon}}\\
        & & 1.3b & 6.7B & 13B & 30B & 66B  & 7B & 13B & 70B &  7B & 40B& 180B \\ \midrule
        FP16 & 0\% & 14.63 & 10.84 & 10.13 & 9.56 & 9.34 & 5.47 & 4.88 & 3.20 &6.59  &5.23 & 3.30  \\  \midrule
        SparseGPT & 2:4 &  24.08  & 14.15  & 12.93  & 10.93 & 10.08 & 10.97  & 8.78  & 5.70  & 12.33  & 12.33 &6.13  \\  \midrule
         
         \multirow{2}{*}{QUIK-8B}   & 0\% &  14.62 & 10.84 & 10.13 &  9.51 & 9.29 & 5.48  &  4.89 & 3.33 &6.59 & 5.23 & 3.31\\  
           & 2:4 &  22.69  & 14.59 &12.87 & 11.06  & 10.24  & 11.07 & 8.66 &  5.89 & 11.07 & 8.09 & 6.19  \\ 
        
    \end{tabular}
    \caption{WikiText2 accuracy results for applying 2:4 sparsity with QUIK-8B. We use 256 outliers in all experiments.}
    \vspace{5mm}
    \label{tab:sparse_int8quantized_results_full}
\end{table*}
\setlength\tabcolsep{5pt}

\section{Falcon performance benchmark}
We also explore the performance improvements of Falcon~\cite{falcon2023} models. The 8xRTX3090 machine contains around 190GB GPU memory which is not enough to run fp16 model inference.

\newpage

\section{Performance on RTX3080 GPUs}
\label{appendix:rtx3080}
\begin{figure}[t!]
\centering
\includegraphics[width=0.35\textwidth]{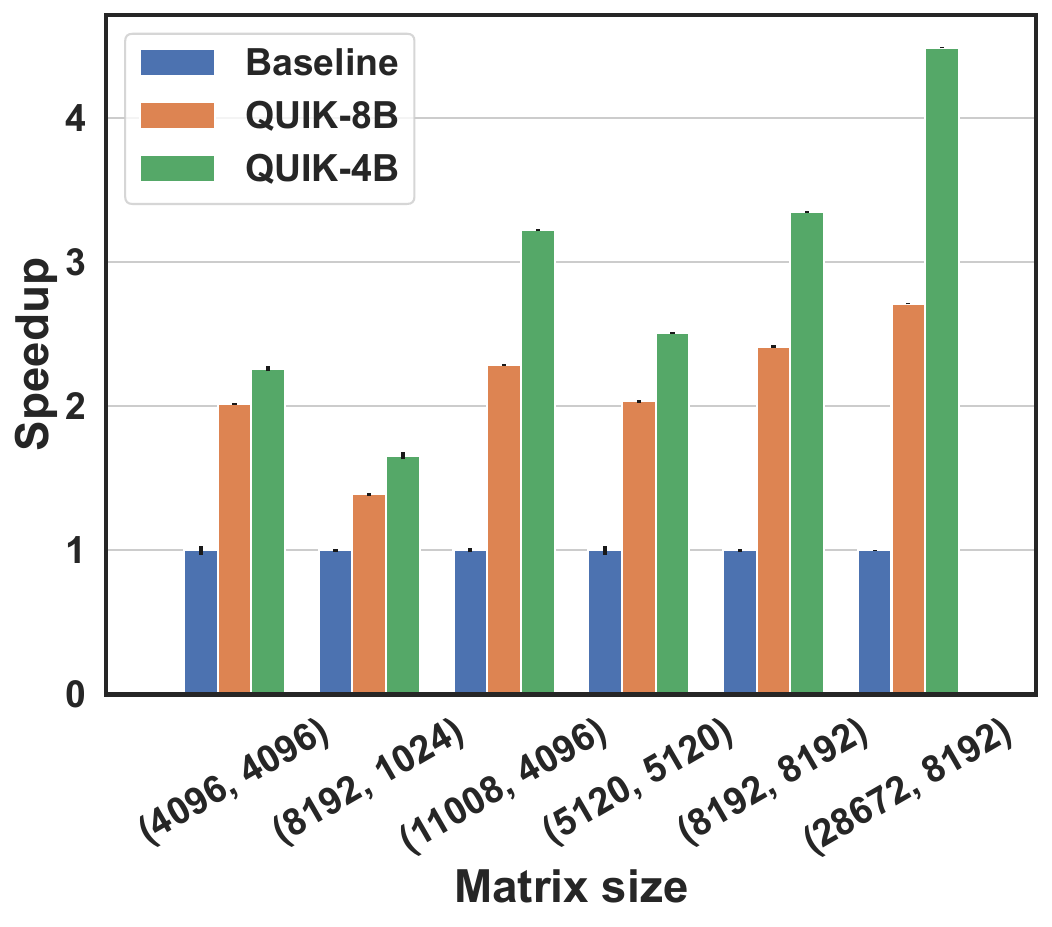}
\caption{\label{fig:bench_layerwise_rtx3080} Layer-wise speedups on a single RTX3080 for different layer sizes and compression types. QUIK-4B with 256 outliers, QUIK-8B without outliers.}
\vspace{-.5em}
\end{figure}

To validate the performance of QUIK in other types of GPUs we conducted benchmarks on RTX3080 GPUs. The results are presented in Figure~\ref{fig:bench_layerwise_rtx3080}. We can see that QUIK-4B still can get more that 4x speedup on another type of GPU.

\section{Performance at different sequence sizes}
\begin{figure*}[h!]
    \centering
    \setkeys{Gin}{width=0.35\linewidth}
    \subfigure[Layerwise Performance.\label{fig:seq_size_layerwise}]{\includegraphics{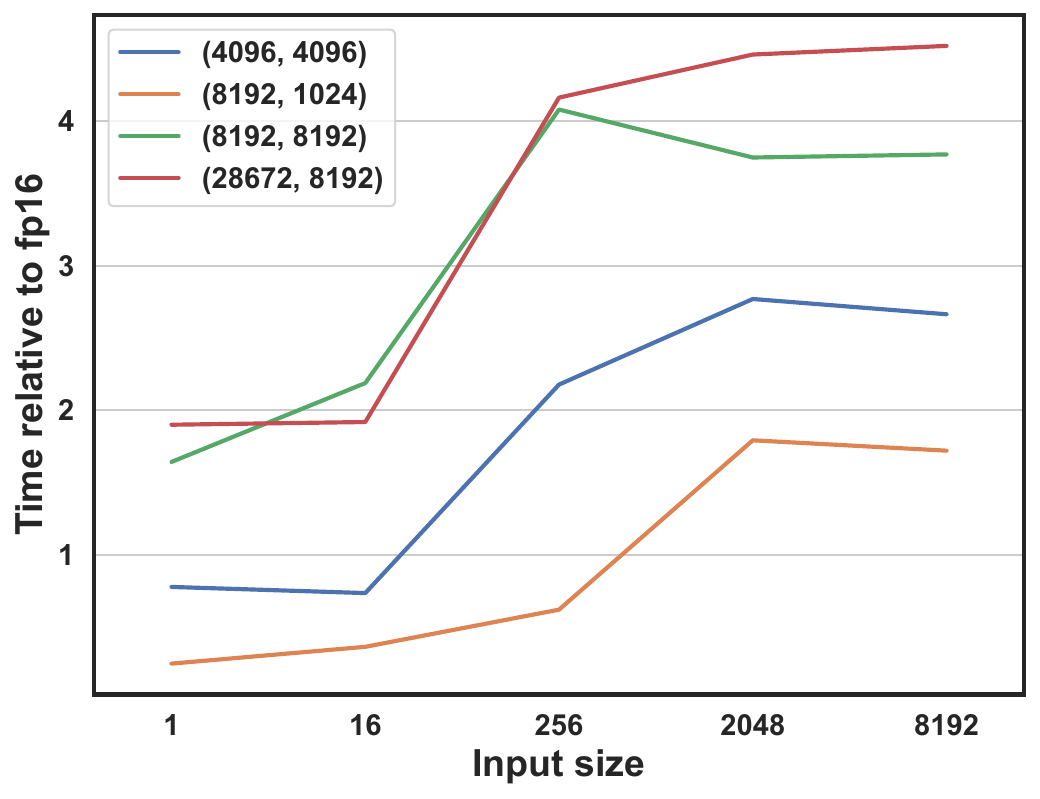}}\hfil
    \subfigure[LLaMA Block performance.\label{fig:seq_size_block}]{\includegraphics{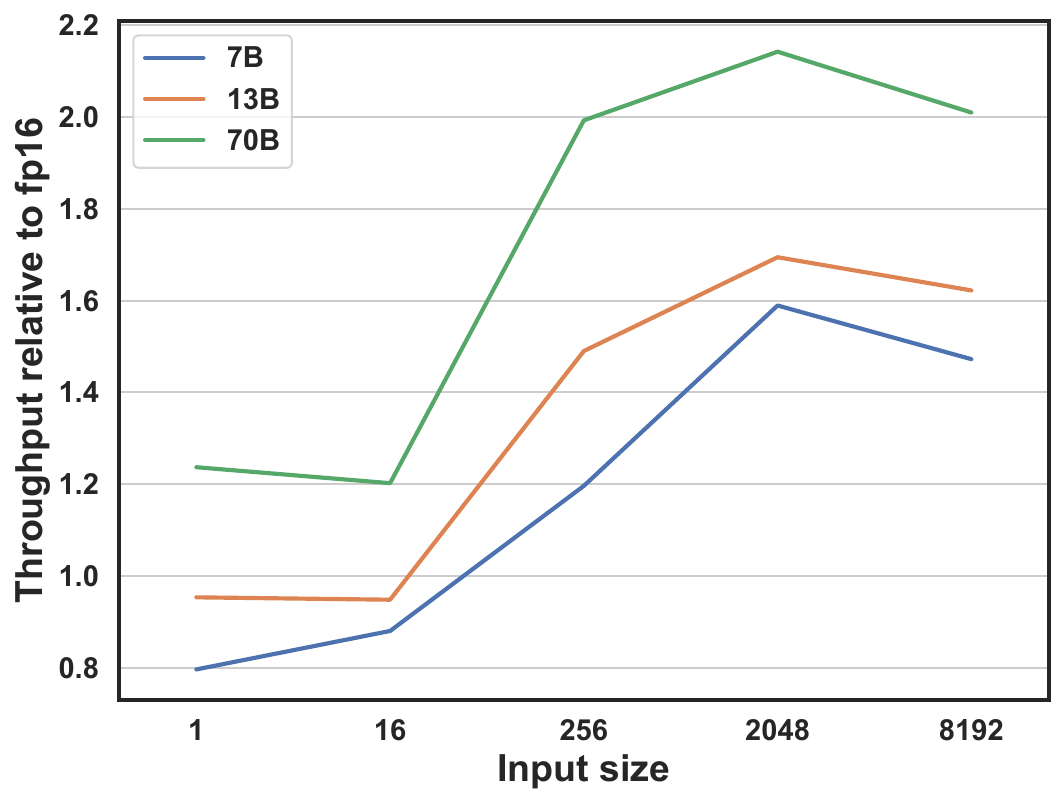}}
    \caption{Relative performance of QUIK-4B with outliers for different sequence sizes (batch size = 1) on RTX3090 GPU}
    \label{fig:seq_size}
\end{figure*}

We mainly focus our work on the ``prefill'' cases with large sequence sizes (in all our experiments sequence size is equal to 2048). In this section we explore the performance of the QUIK-4B with other input sequence sizes. In Figures~\ref{fig:seq_size_layerwise} and~\ref{fig:seq_size_block} we vary input size from 1 to 8k. 
In the first expeeriment (Figure.~\ref{fig:seq_size_layerwise}) we ran layer-wise benchmark, in the second (Figure~\ref{fig:seq_size_block}) we ran inference of a single Transformer block (on a single GPU).
We see that at small input sequence sizes QUIK is noticably slower for smaller layer size and models. It can be explained by the fact that the gains of low precision matrix multiplication at this scale can not compensate the quantization overheads. However, at large layer and model sizes QUIK has up to 2x speedup even with single token input. In case of the large input sequences we see that performance decreases meaning that low precision matrix multiplication saturates at this scale.

\section{Performance with various outlier number}
\begin{figure}[H]
    \centering
    \includegraphics[width=0.45\linewidth]{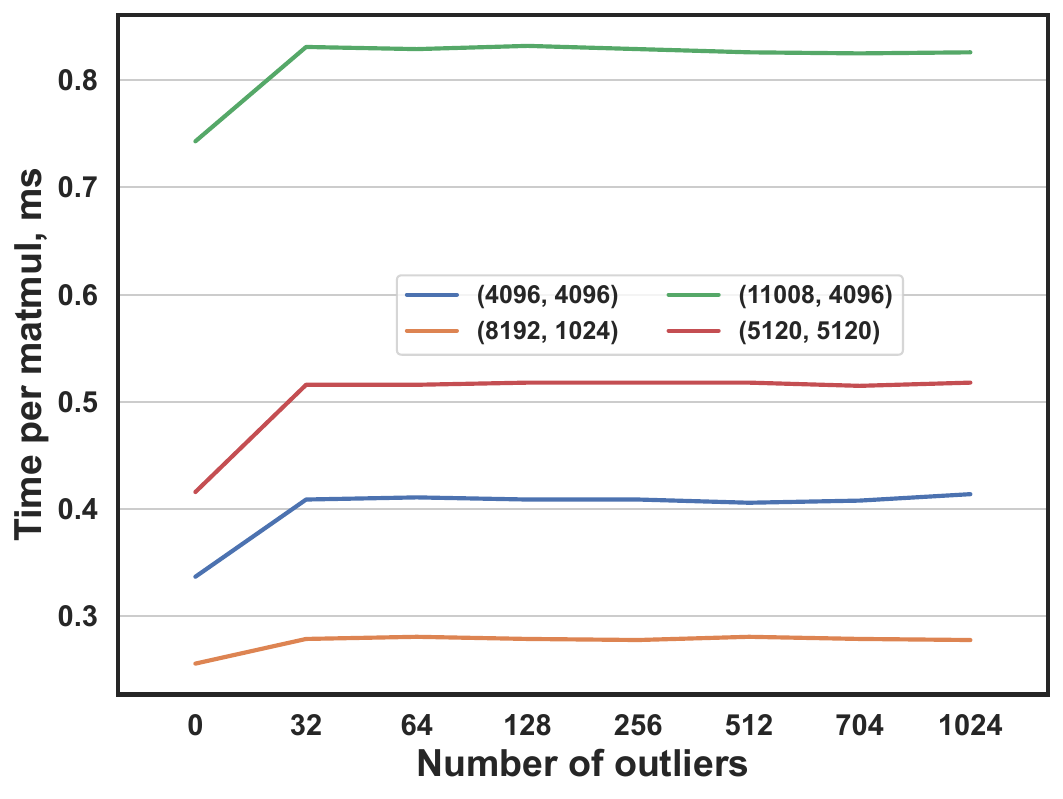}
    \caption{Timing results for different QUIK-4B layers sizes with various number of outliers on RTX3090 GPU.}
    \label{fig:layerwise_outliers}
\end{figure}

In this section we explore the effect of outliers numbers on the QUIK performances. Figure~\ref{fig:layerwise_outliers} suggests that the timing of QUIK matmul stays the same across all layer sizes for all non-zero outlier numbers. The zero outliers case superiority can be explained by the fact that it does not have additional full precision matrix multiplication and input data movements. However, these results show that QUIK allow increase the outlier number without performance sacrifices which is crucial for the accuracy recovery, as we discussed in the Section~\ref{sec:outlier_count}.

